\documentclass{article}
\usepackage{multirow}

\usepackage[accepted]{icml2026}
\usepackage{graphicx}
\usepackage{subcaption}

\usepackage[utf8]{inputenc} %
\usepackage[T1]{fontenc}    %
\usepackage{hyperref}       %
\usepackage{url}            %
\usepackage{booktabs}       %
\usepackage{amsfonts}       %
\usepackage{nicefrac}       %
\usepackage{microtype}      %
\usepackage[dvipsnames]{xcolor}         %
\usepackage{pifont}%
\usepackage{bm}
\usepackage{ml_defs}
\usepackage{multicol}
\usepackage{multirow}
\usepackage{comment}
\usepackage{threeparttable}
\usepackage{enumitem}
\usepackage{listings}
\usepackage{xcolor}
\usepackage{algorithm}
\usepackage{tocbasic}
\usepackage{tocloft}

\newcommand{\xmark}{\ding{55}}%

\definecolor{linkcol}{HTML}{3073AD}  
\definecolor{citecol}{HTML}{3073AD} 
\definecolor{urlcol}{HTML}{3073AD} 

\definecolor{lightblue}{RGB}{32, 194, 217}
\definecolor{jku_red}{RGB}{217, 92, 76}
\definecolor{jku_blue}{RGB}{0, 132, 187}
\definecolor{jku_green}{RGB}{91, 167, 85} 
\definecolor{jku_yellow}{RGB}{241, 188, 63}
\definecolor{jku_cyan}{RGB}{79,176,191}
\definecolor{jku_grey}{RGB}{125,130,140}
\definecolor{jku_lightgreen}{RGB}{191,206,82}
\definecolor{jku_violett}{RGB}{174,97,157}

\usepackage{makecell}

\usepackage{siunitx}

\newcommand{\stoptocwriting}{%
    \addtocontents{toc}{\protect\setcounter{tocdepth}{-5}}}
\newcommand{\resumetocwriting}{%
    \addtocontents{toc}{\protect\setcounter{tocdepth}{2}}}

\hypersetup{
   colorlinks=true,
   linkcolor=linkcol,
   citecolor=citecol,
   urlcolor=urlcol,
}

\begin{document}

\twocolumn[
 \icmltitle{Stabilizing In-Context Multi-Source Domain Adaptation for Biomedical Images Through Controls}

  \icmlsetsymbol{equal}{*}

  \begin{icmlauthorlist}
    \icmlauthor{Ana Sanchez-Fernandez}{jku}
    \icmlauthor{Thomas Pinetz}{muw}
    \icmlauthor{Werner Zellinger}{jku}
    \icmlauthor{Günter Klambauer}{jku,kfi}
  \end{icmlauthorlist}

  \icmlaffiliation{jku}{ELLIS Unit Linz, LIT AI Lab and
  Institute for Machine Learning, JKU Linz, Austria}
  \icmlaffiliation{muw}{Institute of Artificial Intelligence, Center for Medical Data Science, Medical University Vienna}
  \icmlaffiliation{kfi}{Clinical Research Center for Medical AI, JKU Linz, Austria}

  \icmlcorrespondingauthor{Günter Klambauer}{klambauer@ml.jku.at}

  \icmlkeywords{Machine Learning, ICML}

  \vskip 0.3in
]

\printAffiliationsAndNotice{}  %

\stoptocwriting
\begin{abstract}
Biomedical imaging data presents enormous potential for deep learning models to predict invaluable properties, such as diseases and drug effects. However, unavoidable alterations of the technical conditions cause \textit{batch effects}: variations between groups of samples that are not due to any biological signal of interest. Batch effects greatly hinder the generalization abilities of deep learning models, preventing their practical use in the real world. Unsupervised Domain Adaptation (UDA) methods have been proposed to mitigate batch effects, but they usually assume that the data is comprised of only one source domain and one target domain, whereas biological datasets are comprised of multiple domains, both at training and at inference time. While Batch Normalization–based test-time and meta-learning adaptation methods offer a promising mechanism for domain alignment, we show that existing approaches exhibit degraded performance under the  usual inference scenarios of small target batch sizes and label shift. We address these limitations by leveraging negative control samples, which are consistently present in every experimental batch in biological datasets, as stable context for adaptation. We propose CS-ARM-BN, a meta-learning BN adaptation method that uses controls both during training and inference to stabilize domain statistics. We perform a suite of experiments of Mechanism-Of-Action (MoA) classification, a crucial task for drug discovery, on the large JUMP-CP imaging dataset. Our experiments show that CS-ARM-BN substantially improves robustness to batch size and class distribution shifts, enabling practical use of deep learning models for biomedical images.
\end{abstract}

\begin{figure}
    \centering
    \includegraphics[width=0.5\textwidth]{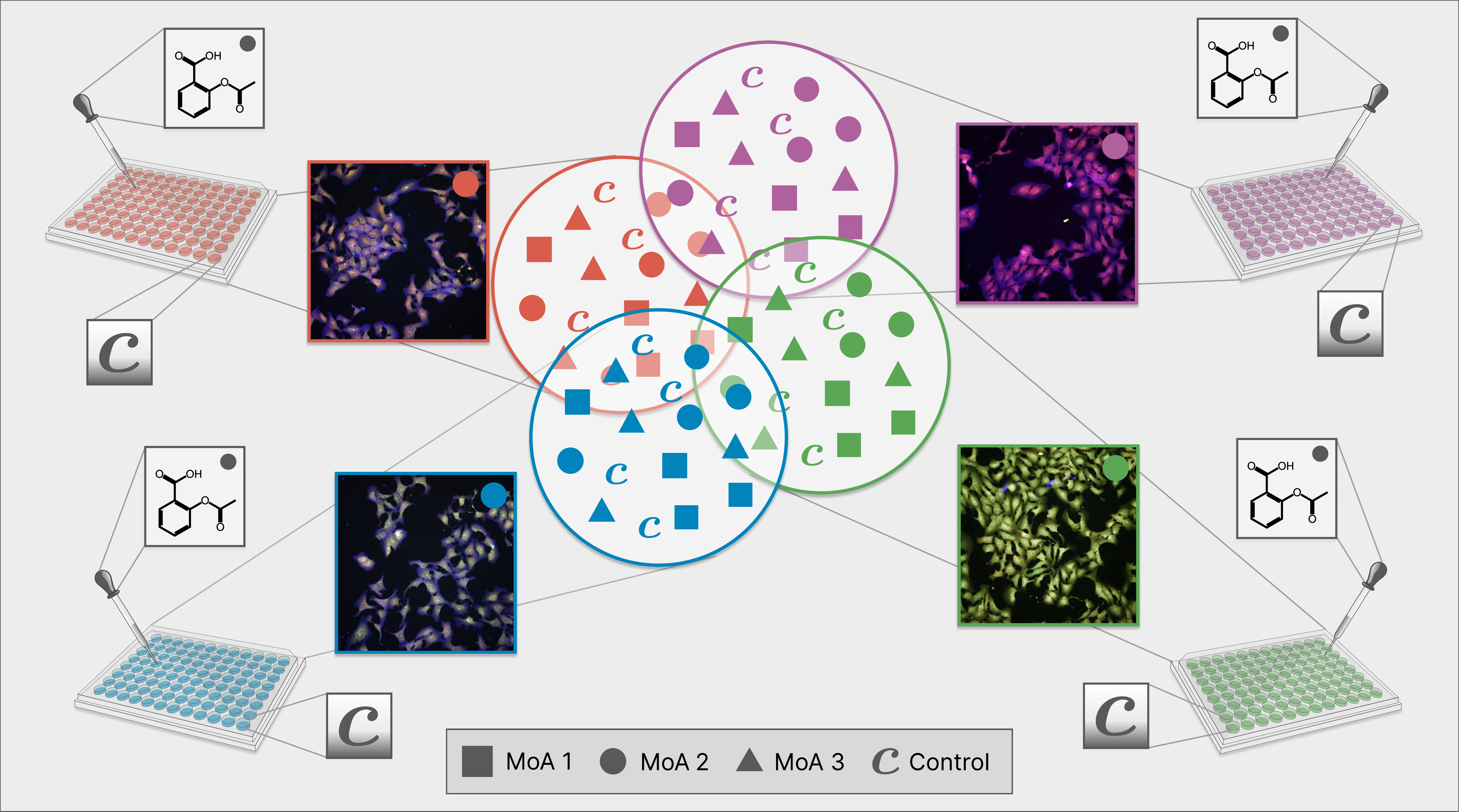}
    \caption{Representation of \textit{batch effects} in microscopy imaging data considered 
    as a multi-source domain adaptation (MSDA) problem. In this setting, each source consists of different experimental conditions, e.g. different plates. In each domain, an image of the same class is depicted, which 
    represents a particular mechanism-of-action (MoA). Control samples are unperturbed samples, that are present in every domain, and which we represent with the $\boldsymbol{c}$ symbol.} 
    \label{fig:general_schema}
\end{figure}

\begin{figure*}
    \centering
    \includegraphics[width=\textwidth]{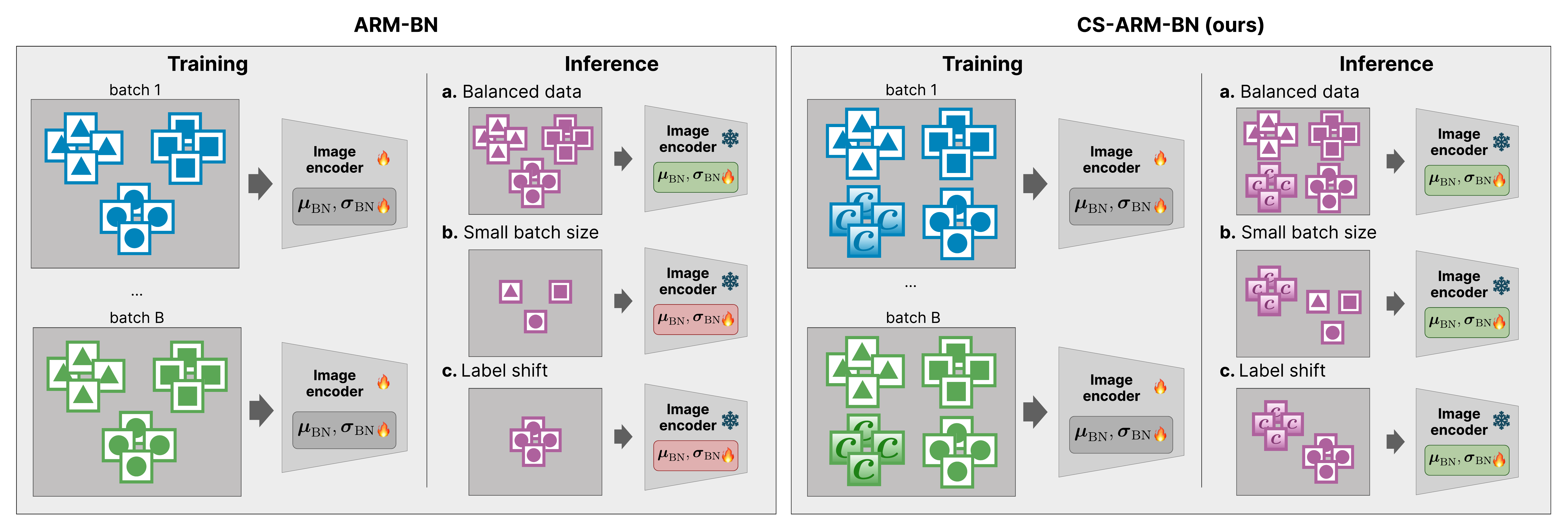}
    \caption{Graphic representation of our method, CS-ARM-BN, and comparison to ARM-BN \citep{Zhang2021}. Both are meta-learning methods that are modified at test-time by using the BN statistics from the target domain (lilac). CS-ARM-BN uses control samples both at training and at inference time, which provides stability when (b) the number of perturbed samples is small or (c) the label distribution is shifted.} 
    \label{fig:method_shema}
\end{figure*}

\section{Introduction}
\label{sec:intro}
\textbf{Batch effects severely degrade model generalization
for biomedical datasets.}
The growing scale of publicly available biomedical 
imaging datasets offers an unprecedented opportunity for 
deep learning models to predict clinically and 
biologically meaningful properties, such as disease 
states, treatment responses, and drug effects 
\citep{Chandrasekaran2023, Zhang2025.02.20.639398, 
Peidli2023}. However, a key challenge arises because 
biological data has to be acquired in \emph{experimental 
batches}, which are groups of samples that were obtained 
under different technical conditions. Even in carefully 
controlled environments, images from different batches 
present variations due to non-biological factors that 
might be more prominent than the biological signal in 
question \citep{Leek2010, Sypetkowski2023, Luecken2021}. 
These variations are called \emph{batch effects} and are 
a well-known artifact, such that unperturbed samples, 
known as \emph{negative controls}, are included in every 
experimental batch as a reference \citep{Bray2017}. 
Negative controls have been long used in the field of 
cellular imaging profiling to try to disentangle 
biological signal from confounding factors, often by 
standardizing features using controls from the 
corresponding experimental batch \citep{Arevalo2024}. 
However, such post-hoc corrections do not allow 
exploiting the capabilities of end-to-end deep learning 
models, and traditional deep learning models perform 
poorly on new batches \citep{Chen2020, Kim2025}.

\textbf{Domain adaptation provides a natural framework 
for batch effect correction.} 
We could consider batch effect correction in machine 
learning as an \emph{unsupervised  domain adaptation} 
(UDA) problem, where the source domain is the training 
data available and the target domain is a new 
experimental batch. Classical UDA aligns source and 
target distributions using approaches such as 
a) importance weighting \citep{shimodaira2000improving, ben2006analysis, kimura2024short}, 
b) distance-based matching \citep{sun2015, Zellinger2017cmd}, 
c) adversarial training \citep{ganin2014}, or 
d) generative translation \citep{NIPS2016_502e4a16, Taigman2016UnsupervisedCI, Bousmalis2017, Chung2024}. 
In the field of microscopy imaging data, test-time 
training (TTT) strategies have been explored 
\citep{Haslum2023}. However, traditional UDA and TTT 
methods assume a small number of source domains, each of 
them with a large amount of samples, and require 
retraining the model every time a new domain needs to be 
evaluated. In contrast, microscopy imaging data is 
comprised of many domains with few samples \citep{Chandrasekaran2023, Bray2017}, 
which makes UDA methods unsuitable in this setting. 
Batch effect correction can be formalized instead 
as a \emph{multi-source domain adaptation} (MSDA) problem (see Figure \ref{fig:general_schema}), 
where each experimental batch is a source domain \citep{Farahani2020, NIPS2011_b571ecea}.

\textbf{Multi-source domain adaptation and in-context test-time adaptation.}
Recent works have been developed to extend classical UDA methods 
to a multi-source setting \citep{zhao2018adversarial, peng2019moment, wen2020domain, yang2020curriculum, venkat2020your, guo2018multi}. Nevertheless, just like UDA, MSDA methods 
require re-training the model every time a new target domain needs to be evaluated. 
In microscopy imaging data, new experimental batches come in sequentially, 
making these methods highly inefficient. 
An ideal setting for microscopy imaging data would be one where the 
multi-source nature of the data can be exploited and also the model 
can be dynamically adapted at test time without need for retraining. 
Adaptive Risk Minimization (ARM) \citep{Zhang2021} is a 
meta-learning framework \citep{hochreiter2001learning, finn2017model} 
that enables such rapid adaptation.

\textbf{Batch Normalization as a domain adaptation mechanism.} 
Batch Normalization (BatchNorm) is a widely established normalization technique, 
that allowed training deep neural networks 
with skip connections \citep{Sergei2015}, and 
that can additionally be used as a module for domain adaptation \citep{Li2016}. 
Conventionally, when the assumption that both training and test data 
are sampled from the same distribution is made, 
the stored training statistics are used at inference time. 
In contrast, when the test samples are known to come from a different distribution, 
adapting these layers by using the statistics of the new domain 
at inference time provides a straight-forward 
test-time domain adaptation method \citep{Li2016, wang2021tent}. 
However, it has also been shown that these methods fail 
when not enough samples are available to accurately estimate 
the new domain statistics or when the class distribution 
is shifted at test time \citep{Zhao2023, Park2023}.

\textbf{Control samples provide an opportunity to stabilize the context in meta-learning methods.} 
In addition to the methods presented above, there are also meta-learning methods 
which include adapting the batch normalization layers in a meta-learning setting \citep{Zhang2021}. 
The main difference between BatchNorm adaptation \citep{Li2016} and 
meta-learning BatchNorm adaptation \citep{Zhang2021} is that 
the adaptation does not only occur at test-time but also during training, 
such that the model learns to adapt to new batches. 
This has been shown to provide a significant improvement with 
respect to its pure test-time counterparts. Nevertheless, in this work, 
we show that these methods also underperform in the presence of 
small batch sizes and under label shifts.   
Both scenarios, a) small sample sizes and b) label shifts in a new experimental batch 
are very plausible, because of the typical design of such bioimaging studies \citep{Hughes2011, Knowles2003}. 
However, almost all biomedical experiments guarantee that a specific type of 
samples will be \emph{present in every new experimental batch (i.e., domain}): 
\emph{negative controls} \citep{Chandrasekaran2023, Bray2017}. 
Therefore, we propose to exploit the presence of negative controls and 
use them as context samples both during training and for inference 
of a meta-learning model, showing that this approach makes the performance 
of our method robust to small numbers of new experimental batches and
to shifts in the class distribution.
\vspace{-5pt}
\paragraph{Contributions.}
Our main contributions are the following:
\begin{itemize}[itemsep=0pt, topsep=0pt]
    \item We propose \textbf{CS-ARM-BN}, a control-stabilized meta-learned 
    BatchNorm adaptation method that leverages negative control samples 
    available in every experimental batch, and show that it robustly 
    alleviates the failure modes of previous methods and nearly closes 
    the generalization gap between in-domain and unseen-batch performance.

    \item We demonstrate that adaptive Batch Normalization--based methods, 
    particularly when combined with meta-learning and in-context adaptation, 
    provide an effective and computationally efficient approach to mitigate
    severe batch effects in biomedical microscopy images.
    
    \item We systematically analyze and expose key limitations 
    of existing BatchNorm-based adaptation methods, 
    including AdaBN and ARM-BN, showing that their performance degrades in realistic settings such as label shift.
    
\end{itemize}

\section{Problem setting and Background}
We consider the problem of adapting from $B$ source batches of size $N$ to a single target batch of size $M$ under the cross-entropy loss $\mathcal L_{\mathrm{CE}}(\By,\hat \By )=- \sum_{k=1}^K y_k \log(\hat y_k)$. This problem can be extended to any loss function $\mathcal L$, many target batches and different batch sizes, in a straightforward way.

We are given a source dataset $\mathcal D^{\mathrm{train}}=\left\{\mathcal{D}^{(1)}, \ldots, \mathcal{D}^{(B)}\right\}\subset \left(\mathcal{X}\times\mathcal{Y}\right)^N$ consisting of $B$ experimental batches $\mathcal{D}^{(b)}=(\Bx^{(b)}_n, \By^{(b)}_n)_{n=1}^N$ drawn from distributions $p_{xy}^b, b\in\{1,\ldots,B\}$, respectively.
Following~\citet{baxter1998theoretical,Zhang2021}, we assume that the batches are independent and drawn from a meta-distribution $\mu$.

The goal is to find a prediction model $\Bg(\Bx,\Bw)=\hat \By$ together with an adaptation method $\mathrm{Adapt}\!\left(\Bw,\Bg,\mathcal{D}^\beta\right)=\widetilde{\Bw}$ that allow us, for any new \textit{unlabeled} experimental target batch $\mathcal{D}^\beta=(\Bx_n^\beta)_{n=1}^M$, to adapt the weights $\Bw$ to new weights $\widetilde{\Bw}$ with a small target batch risk
\begin{align}
    \frac{1}{M}\sum_{m=1}^M \mathcal \ell \big( \By_m, \Bg \left(\Bx_m, \widetilde{\Bw}\right) \big),
\end{align}
whenever the target batch is drawn from the marginal $p_x^\beta$ of a new target distribution from $\mu$. $\ell$ is an appropriate loss function.

\textbf{Feature extractor and classifier networks.} 
We denote the following components that are 
normally used by domain adaptation methods:
a) \emph{an encoder or feature extractor} network $\Bf: \mathcal X \mapsto \mathbb{R}^{D}$ or $\Bf_{\Bw}$ with
weights $\Bw$,
which maps the input images $\Bx \in \mathcal X$ 
to an embedding $\Ba \in \mathbb{R}^{D}$, and
b) \emph{a classifier} $\Bc: \mathbb{R}^{D} \mapsto \mathcal Y$,
which maps the embeddings to a class label $\hat \By \in \mathcal Y$ in the output space $\mathcal Y$.

The model $\Bg: \mathcal X \mapsto \mathcal Y$ 
is usually a composition of the feature extractor
with the classifier
$\Bg = \Bc \circ \Bf: \mathcal X \mapsto 
\mathcal Y$.

\subsection{Domain Adaptation Methods for Biomedical Images}
\label{subsec:others}
We first introduce the main domain adaptation methods used for comparison and referenced throughout the manuscript. See Supplementary Section~\ref{appsec:methods} for descriptions of all compared methods.

\textbf{Supervised baselines. ResNet~\citep{He2015} and Foundation Model CA-MAE~\citep{Kraus2024}.}
We train a standard classifier $\Bg$ on $\mathcal D^{\mathrm{train}}$ by minimizing
\begin{align}
\min_{\Bw} \sum_{b=1}^B \;\sum_{(\Bx,\By)\in \mathcal D^{(b)}} \mathcal L_{\mathrm{CE}}(\By, \Bg(\Bx,\Bw))
\end{align} with cross-entropy loss $\mathcal L_{\mathrm{CE}}$ and 
using a ResNet50 architecture with either BatchNorm (BN) or InstanceNorm (IN) layers, and a linear layer as a classifier.
This classifier is not adapted to new experimental batches.

For the foundation model baseline, we use a frozen CA-MAE encoder $\Bf_{\mathrm{MAE}}$ to obtain embeddings $\Ba=\Bf_{\mathrm{MAE}}(\Bx)$ and perform linear probing with a classifier $\Bc_{\theta}$ trained with CE-loss.

\textbf{Adaptive Batch Normalization (AdaBN) \citep{Li2016}.} In Adaptive Batch Normalization, the adaptation of the model is simply carried out by using the BN statistics of the target domain at inference time, instead of those of the training data. For microscopy images \citet{Lin2022} proposed adapting the BN statistics
based on control samples. 

\textbf{Fully Test-Time Adaptation by Entropy Minimization (TENT) \citep{wang2021tent}.} In TENT, the adaptation is also carried out at test-time, and, as in AdaBN, the BN statistics are computed using samples from the new domain. However, in this method, an additional adaptation is included, consisting of minimizing the entropy $H$ of the model predictions $\hat \By$ by updating only the BatchNorm affine parameters $\kappa$ and $\lambda$ :
\vspace{-8pt}
\begin{align}
    \hat \By=\Bg(\Bx), \qquad 
    H(\hat \By) = - \sum_{k=1}^K \hat y_{k} \log(\hat y_{k})
\end{align}
\vspace{-5pt}
\begin{align}
    \min_{\kappa, \lambda} H(\hat \By)
\end{align}

\vspace{-5pt}

\textbf{Adaptive Risk Minimization (ARM) / In-context adaptation~\citep{Zhang2021}.}

ARM uses a meta-learning objective \citep{hochreiter2001learning,finn2017model}:
\begin{align}
\min_{\Bw, \Bphi} \;
\frac{1}{B} \sum_{b=1}^B \;
\frac{1}{M} \sum_{(\Bx_m,\By_m) \in \mathcal D_M^b}
\mathcal L_{\mathrm{CE}} \big(\By_m, \Bg(\Bx_m ; \Bh_{\Bphi}(\Bw,\Bx^{b}_{1:M}) \big),
\end{align}

where $\mathcal D_M^b$ is a random subset of $M<N$ pairs of
batch $b$, considered as \emph{context set}. 
$M$ is usually in the size of the
expected future dataset from the new domain, in our case
the size of an experimental batch.

The central part is the test-time adaptation function 
\begin{align}
\Bh_{\Bphi}: \; \mathcal W \times \mathcal X^M \to \mathcal W,\qquad 
\Bw^{\beta} = \Bh_{\Bphi} \big(\Bw, \Bx^{\beta}_{1:M}\big),
\end{align}
which maps the current model parameters $\Bw$ and \emph{context} (the target-batch samples $\Bx^{\beta}_{1:M}$) to the adapted parameters $\Bw^{\beta}$. The test-time adaptation function 
$\Bh_{\Bphi}$ has itself parameters $\Bphi$, which are learned 
with the usual gradient descent techniques.
At test time on a new experimental 
batch $\beta$, we first compute $\Bw^{\beta}=\Bh_{\Bphi}(\Bw,\Bx^{\beta}_{1:M})$ and then predict with $\hat \By=\Bg(\Bx;\Bw^{\beta})$ for any $\Bx$ from batch $\beta$.

\emph{ARM-BN.}
In this variant of ARM, the adaptation function $\Bh_{\Bphi}$  only
replaces the batch normalization statistics
$\Bmu_{\mathrm{BN}}, \Bsigma_{\mathrm{BN}}$ 
of all batchnorm-layers of the main network $\Bg$
with the statistics from the batch $\beta$ from 
the new domain. 

\emph{ARM-CML.}
In this version, the adaptation function $\Bh_{\Bphi}$ uses the unlabeled samples from the new batch to compute a compact representation, 
a \emph{context vector}, 
that captures the specific characteristics of the target domain. 
Unlike ARM-BN, which only updates normalization statistics, ARM-CML performs a lightweight parameter adaptation conditioned on the new data, enabling the model to better align with the target domain without requiring labels or retraining.

\section{CS-ARM-BN: Control-Stabilized Adaptive Risk Minimization with Batch Normalization}

A key limitation of BN-based test-time and meta-learned adaptation methods is their reliance on statistics estimated from a set of \emph{unlabeled target samples}. When the number of available samples is small or when the class distribution at test time differs from that observed during training, the resulting batch normalization (BN) statistics become noisy and confounded by label shift \citep{Park2023, Zhao2023}, leading to unstable or degraded performance. We confirm this occurrence in our experiments shown in Sections \ref{sec:exp3_label_shift} and \ref{sec:exp4_label_shift_new_source} and Appendix Sections \ref{appsec:exp3_batch_size} and \ref{appsec:label_shift}.

In high-content screening experiments, and generally in biomedical experiments, \emph{negative control samples} are available \citep{Chandrasekaran2023,Arevalo2024}. 
For every experimental batch $\beta$, 
a set of control images $\{\Bz_c^{\beta}\}_{c=1}^{C}$
is acquired under the same technical conditions as the perturbed samples, but without inducing any biological effect. These controls provide a direct estimate of batch-specific technical variation that is independent of the class composition of the perturbed samples.

\paragraph{Control-stabilized adaptation.}
In CS-ARM-BN, we extend ARM-BN by incorporating negative control samples into the context used for adaptation. Instead of estimating BatchNorm (BN) statistics solely from the unlabeled perturbed samples $\Bx^{\beta}_{1:L}$, we compute the adaptation statistics using the union of control and perturbed samples from the target batch,
\[
\mathcal C^{\beta} \;=\; \{\Bz^{\beta}_c\}_{c=1}^{C} \;\cup\; \{\Bx^{\beta}_l\}_{l=1}^{L}.
\]
Concretely, for each BN layer, the adapted mean and variance are computed over this combined context set as
\begin{align}
\Bmu_{\mathrm{BN}}^{\beta} 
&= \frac{1}{|\mathcal C^{\beta}|}
\sum_{\Bu \in \mathcal C^{\beta}} \Bu, \\
(\Bsigma_{\mathrm{BN}}^{\beta})^2 
&= \frac{1}{|\mathcal C^{\beta}|}
\sum_{\Bu \in \mathcal C^{\beta}}
\big(\Bu - \Bmu_{\mathrm{BN}}^{\beta}\big)^2,
\end{align}
where $\Bu \in \mathcal C^{\beta}$ denotes the BN-layer activations corresponding to either a control image $\Bz^{\beta}_c$ or a perturbed image $\Bx^{\beta}_m$.
These adapted statistics replace the running training statistics in all BN layers of the prediction model $\Bg$ when evaluating samples from batch $\beta$.

Another option to mitigate the effects of label shift and small batch sizes could be to use only control samples to adapt the Batch Normalization statistics, as proposed by Batch Effect Normalization (BEN) 
\citep{Lin2022}. However, in the following paragraph, we argue that including both perturbed and control 
samples provides the best trade-off between bias and variance. Empirically, we found that adapting BN layers using control samples only at evaluation time performed poorly, but when only controls were already used during training, the performance improved (see Table~\ref{tab:label_shift_new_source}). We evaluate BEN and its meta-learned variant ARM-BEN in our most challenging setting (Section~\ref{sec:exp4_label_shift_new_source}), which combines severe label shift with cross-source transfer. This is the regime in which the bias-variance trade-off between the three estimators (see below)
distinguishes them the most. Under mild label shift, the bias that BEN seeks to avoid by accepting a higher variance is small to begin with, leaving BEN dominated by ARM-BN, which is an outcome the bias–variance analysis already predicts.

Let $M = C + L$ be the total number of samples in a new target
batch, where 
$C$ is the number of control samples
and $L$ is the number of non-control or perturbed samples.
We model the BN activation mean for sample $x$ as:
\vspace{-1pt}
$$\mu_{\text{obs},x} = \mu_{\text{domain}} + \mu_{\text{class}(x)}$$
\vspace{-1pt}
where $\mu_{\text{domain}}$ 
is the batch-specific technical offset we wish to estimate, and
$\mu_{\text{class}(x)} = 0$ for controls and 
$\mu_{\text{class}(x)} \neq 0$ for perturbed samples.

The three estimators of $\mu_{\text{domain}}$ and their mean 
squared errors (MSEs) are:

\begin{enumerate}
\item \textbf{ARM-BN or AdaBN estimator (perturbed samples only):}
\begin{align}
\text{MSE}(\hat{\mu}^{\text{ARM-BN}}) = \underbrace{\bar{\mu}
_{\text{class}}^2}_{\large\text{large bias}^2} + \frac{\sigma^2}{L}
\end{align}
The bias $\bar{\mu}_{\text{class}}$ grows with the degree of label shift, explaining the
performance collapse observed in Tables~\ref{tab:label_shift} and \ref{tab:label_shift_new_source}  .
\vspace{-1pt}
\item \textbf{Control-based estimator (controls only):}
\begin{align}
\text{MSE}(\hat{\mu}^{\text{ctrl}}) = \underbrace{0}_{\large\text{no bias}} + \frac{\sigma^2}{C}
\end{align}
The estimator is unbiased, but 
suffers from high variance when $C$, the number of controls,
is small.
\vspace{-1pt}
\item \textbf{Our estimator CS-ARM-BN (controls + perturbed samples):}
\begin{align}
\text{MSE}(\hat{\mu}^{\text{CS-ARM-BN}}) = \underbrace{\left(\frac{L}{M}\right)^2 \bar{\mu}_{\text{class}}^2}_{\large\text{small bias}^2} + \frac{\sigma^2}{M}
\end{align}
exploits all available samples $M=C+L$ and allows for a small bias, but decreases variance by a factor of $1/M$.
\end{enumerate}
\vspace{-1pt}
In typical settings, 
CS-ARM-BN achieves the lowest MSE 
by trading a small residual bias, shrunk by the
factor $(L/M)^2$ relative to ARM-BN,
against a reduced variance, since
$M > C$. 
This bias-variance tradeoff is most favorable precisely in the
operating regime of real drug screening: 
availability of few controls ($C$)
and moderate-to-severe label shift, 
which is confirmed empirically in Table~\ref{tab:label_shift_new_source}. Details in Appendix~\ref{sec:details-estimators}.

\paragraph{Relation to ARM.}
Importantly, our CS-ARM-BN method 
remains fully within the Adaptive Risk Minimization (ARM) framework. As shown in Figure \ref{fig:method_shema}, during training, the model is meta-learned episodically: for each source batch $b$, control samples and perturbed samples are used as context to adapt BN statistics, and performance is optimized on perturbed query samples from the same batch. This trains the base network to \emph{expect} and \emph{exploit} control-conditioned normalization at test time, rather than treating it as a heuristic post-hoc correction.
\vspace{-1pt}

Compared to standard ARM-BN and other domain adaptation methods, CS-ARM-BN offers three key advantages:
\begin{itemize}
    \item \textbf{Stability under small context sizes:} since all samples, perturbed plus control samples are used in each batch,
    the BN statistics can be estimated reliably even when only few perturbed samples are available.
    \item \textbf{Robustness to label shift:} control samples are independent of the class distribution in the new domain,
    and are used to decrease the bias in the estimator at test time.
    \item \textbf{No additional supervision or re-training:} as in ARM-BN, adaptation remains fully unlabeled and lightweight, requiring only a single forward pass over the samples from the
    new domain 
    i.e. experimental batch, 
    and no expensive test-time training.
\end{itemize}
\vspace{-10pt}

\begin{table*}[h]
\centering
\caption{MoA prediction accuracy on \textbf{in-domain} and 
\textbf{unseen experimental batches} from a \textbf{new domain}.
Column "\textbf{TTA}" 
indicates that the method can adapt itself
given samples from a new domain
at test time, i.e. test time adaptation. 
“\textbf{In-context}” means that the model adapts its parameters based on the test samples through in-context learning. The accuracy values denote mean and standard deviation over 5-fold cross-validation, thus five training re-runs of each method. "FM" indicates a foundation model. 
\label{tab:results}}
\resizebox{\linewidth}{!}{
\begin{threeparttable}
\begin{tabular}{llcccc}
\toprule
\textbf{Method} & \textbf{Method}  & \textbf{In-domain} & \textbf{New experimental }  & \textbf{TTA}  & \textbf{In-context}  \\[-0.25em] 
\textbf{category} & & \textbf{accuracy} & \textbf{batch accuracy} &  & \\ 
\midrule
\multirow{4}{*}{Baselines} 
& no adaptation (BatchNorm)   & 0.939 ± 0.005 & 0.862 ± 0.060 & \xmark & \xmark \\
& no adaptation (InstanceNorm) &  0.949 ± 0.007 & 0.891 ± 0.043 & \xmark & \xmark \\ 

& CA-MAE (FM) \citep{Kraus2024}       & 0.747 ± 0.006
 & 0.709 ± 0.041 & \xmark & \xmark \\
& CellProfiler \citep{Stirling2021}     & 0.923 ± 0.031 & 0.860 ± 0.032 & \xmark & \xmark \\
\midrule
Adversarial & DANN \citep{ganin2014}           & - & 0.911 ± 0.020 & \xmark & \xmark\\ 
\midrule
\multirow{6}{*}{Distance} 
& CORAL \citep{sun2015}           & – & 0.893 ± 0.054 & \xmark & \xmark \\ 
& CA-MAE (FM) + TVN \citep{Ando2017}  & – & 0.747 ± 0.013 & \checkmark & \xmark \\
& TENT \citep{wang2021tent}   & – & 0.929 ± 0.013\tnote{a}  & \checkmark & \xmark \\
& DEYO \citep{Lee2024}   & – & 0.929 ± 0.013\tnote{a} & \checkmark & \xmark \\
& ROID \citep{Marsden2023}  & - & 0.929 ± 0.013\tnote{a} & \checkmark & \xmark \\
& AdaBN \citep{Li2016} & – & 0.928 ± 0.013 & \checkmark & \xmark \\

\midrule
Generative  & Style-ID \citep{Chung2024}           & – & 0.781 ± 0.015 & \checkmark & \xmark\\ 
\midrule
\multirow{3}{*}{Meta-learning} 
& ARM-CML \citep{Zhang2021} & – & 0.843 ± 0.074 & \checkmark & \checkmark \\
& ARM-BN \citep{Zhang2021}  & – & \textbf{0.935 ± 0.018} & \checkmark & \checkmark \\
& CS-ARM-BN (ours)  & – & \textit{0.930 ± 0.019} & \checkmark & \checkmark \\
\bottomrule
\end{tabular}
\begin{tablenotes}
  \small
  \item[a] In this mild-shift, label-balanced regime, the selection and weighting mechanisms of DEYO and ROID practically reduce to TENT's plain entropy update on the BatchNorm affine parameters; accuracies differ only at the fourth decimal.
\end{tablenotes}
\end{threeparttable}
}
\vspace{-11pt}
\end{table*}

\section{Related work}
\textbf{Traditional batch effect correction methods.} Batch effect correction methods that are based on either linear transformations \citep{Johnson2007, Ando2017}, nearest neighbors \citep{Haghverdi2018, Stuart2019, Hie2019} or a mixture of PCA and clustering \cite{Korsunsky2019} have been previously proposed. One of the closest related works 
is \citet{Arevalo2024}, who investigate batch correction 
methods for handcrafted features for microscopy images.
These methods, however, have been originally designed to correct batch effects in feature vectors and cannot be used together with a model trained end-to-end directly from images.  

\textbf{Deep learning and batch effects in microscopy imaging data.}
While \citet{Lin2022} also proposed using negative controls to estimate BN statistics, their work is not framed as meta-learning, and they do not show the scenarios where control-based statistics are beneficial in comparison to using only perturbed samples.
\citet{Palma2025, Lopez2018, Chung2024} presented generative models that could be used to address batch effects. Due to computational constraints, we use \citet{Chung2024} as representative for this class of methods.

\textbf{In-context learning for biological data.} Recently, in the field of single cell transcriptomic data, \citet{Dong2026.01.09.698608} introduces a foundation model that can perform in-context predictions from unlabeled cells. However, their work focuses on tabular data and the transformer architecture.  

\textbf{Test-time adaptation and robustness to label shift.} Aside from the test-time adaptation methods compared in this work, additional TTA methods are available. \citet{Wang2022, Gong2022, Boudiaf2022} propose methods to overcome catastrophic forgetting in online TTA, while our proposed method does not consist on continually adapting the model weights. \citet{Park2023} introduced a hypernetwork to address label shift, but it needs additional training of the model and it is not specifically tailored for biological data where control samples are available.

\section{Experiments and Results}
\label{sec:experiments}

We performed a set of experiments to evaluate the effectiveness of a wide selection of methods for addressing batch effects under different scenarios: \textbf{a) Mild domain shift: new experimental batches.} We evaluate adaptation performance when the target data originate from previously unseen batches generated under similar experimental conditions, reflecting routine batch-to-batch variability. \textbf{b) Strong domain shift: transfer across sources.} We consider a more challenging setting in which the target data come from different institutions or data-generating sources, inducing substantial domain shifts due to variations in protocols, equipment, and experimental practices. \textbf{c) Realistic scenario 1: new experimental batch with label
shifts.} We study the impact of changes in class proportions, evaluating how well methods maintain performance when label distributions differ at inference time. This is the most realistic scenario, since novel sets of compounds are usually screened, which do not maintain the label distribution of the training data. One obvious example is drug discovery efforts, where the compounds that are tested in the lab are suspected to be active towards one specific protein target. \textbf{d) Realistic scenario 2: new imaging source with label shifts.} Finally, we evaluate the performance of our method and the baselines in a setting where test samples exhibit both class distribution shift and domain shift, but are additionally acquired from a new imaging source, yielding a more challenging scenario than the previous one.

\begin{table*}[t]
\centering
\caption{Comparison of MoA classification accuracy of different methods across target sources. The element before the arrow indicates the training domain, the element after the arrow indicates the evaluation domain. The accuracy values denote the mean and standard deviation over 3 re-runs with different seeds.}
\label{tab:generalization_full}
\begin{threeparttable}
\begin{tabular}{l cccc}
\toprule
\multirow{2}{*}{\textbf{Method}} &
\multicolumn{4}{c}{\textbf{Source $\rightarrow$ Target}} \\
\cmidrule(lr){2-5}
 & \textbf{S3 $\rightarrow$ S3} & \textbf{S3 $\rightarrow$ S8}
 & \textbf{S8 $\rightarrow$ S8} & \textbf{S8 $\rightarrow$ S3} \\
\midrule

no adaptation (BatchNorm)
 & 0.939 ± 0.005 & 0.670±0.016 & 0.974 ± 0.002 & 0.498±0.011 \\
no adaptation (InstanceNorm)
 & 0.949 ± 0.007 & 0.796±0.018 & 0.984 ± 0.001 & 0.552±0.014 \\
CA-MAE (FM)
 & 0.747 ± 0.006 & 0.389±0.015 & 0.840 ± 0.004 & 0.356±0.011 \\
CellProfiler
 & 0.923 ± 0.031 & 0.430±0.074 & 0.991 ± 0.004 & 0.550±0.029\\
 \midrule
DANN
 & -- & 0.639±0.175 & -- & 0.324±0.079 \\
\midrule
CORAL
 & -- & 0.804±0.018 & -- & 0.697±0.010 \\
CA-MAE (FM) + TVN
 & -- & 0.688±0.005 & -- & 0.488±0.010 \\
TENT
 & -- & 0.835±0.033 & -- & 0.785±0.008 \\
DEYO
 & -- & 0.836±0.033 & -- & 0.784±0.007 \\
ROID 
 & -- & 0.835±0.034 & -- & \textit{0.788±0.010} \\
AdaBN
 & -- & 0.838±0.037 & -- & \textit{0.788±0.003} \\
\midrule
StyleID
 & -- & 0.774±0.004 & -- & 0.713±0.016 \\
\midrule
ARM-CML
 & -- & 0.645±0.034 & -- & 0.569±0.030 \\
ARM-BN
 & -- & \textit{0.874±0.008} & -- & \textbf{0.795±0.002} \\
CS-ARM-BN (ours)
 & -- & \textbf{0.884±0.005} & -- & 0.776±0.016 \\
\bottomrule
\end{tabular}

\end{threeparttable}
\end{table*}

\textbf{Dataset.} We conduct experiments on the large JUMP-CP dataset \citep{Chandrasekaran2023}. We use source 3, and source 8 in Exp. \ref{sec:exp2_strong_shift}, and perform a classification task to predict eight well-defined MoAs. Therefore, we work with samples perturbed with eight different compounds, each one of them known to produce one of these MoAs (see Appendix Table~\ref{tab:cpds}), and which are present in every plate of the dataset. Each microscopy image has five channels; we use the preprocessing described in Appendix Section \ref{app:data}.

\textbf{Data splits.} To evaluate robustness to batch effects, we consider two data splitting strategies reflecting different levels of domain shift (see Appendix Section \ref{appsec:micro_structure}). In most experiments, we adopt an \emph{experimental-batch split}, where training and evaluation are performed on disjoint experimental batches, representing a challenging but realistic batch generalization setting. For the strong domain shift scenario, we instead evaluate on images acquired from a completely unseen imaging \emph{source}, corresponding to a cross-platform or cross-institution transfer setting. Under the experimental-batch split, we use \textbf{5-fold cross-validation} and report \textbf{mean accuracy $\pm$ standard deviation} across folds, corresponding to five independent training runs per method. Unless otherwise specified, accuracy is reported at the image level.

\textbf{Methods and implementation.} We compare supervised baselines (ResNet50 with BN/IN; CA-MAE linear probe; CellProfiler features), classical UDA/MSDA baselines (DANN, CORAL), generative style transfer (Style-ID), and test-time adaptation / in-context methods (AdaBN, TENT, DEYO, ROID, ARM-CML, ARM-BN). Training details and hyperparameter search spaces are provided in Appendix Table~\ref{tab:hyperparams}. For methods that require \emph{test-time} samples to adapt (AdaBN, TENT, ARM-BN, ARM-CML), we evaluate them in a strictly \textbf{unlabeled} test-time adaptation setting unless explicitly stated.

\subsection{Mild domain shift: new experimental batches within a source}
\label{sec:exp1_mild_shift}
We first study the most common practical scenario: training and test data originate from the \emph{same} imaging source, but the test batches correspond to \emph{new experimental batches}. This setting reflects \emph{mild} domain shift: acquisition conditions drift across experiments, but the imaging platform remains unchanged. Results are summarized in Table~\ref{tab:results}.

\textbf{Batch effects degrade generalization performance.}
A standard ResNet50 achieves high in-domain accuracy (e.g., $0.939 \pm 0.005$ with BN), but drops substantially when evaluated on unseen experimental batches ($0.862 \pm 0.060$). This gap quantifies the severity of batch effects even under mild shifts.

\textbf{Classical UDA baselines are insufficient or impractical.}
Domain-invariant learning (DANN) and discrepancy/moment matching (CORAL) do not reliably close the gap in this few-shot multi-source regime. Moreover, these approaches conceptually assume access to the target distribution during training and are thus inefficient in sequential settings where new batches continually arrive.

\textbf{Test-time and in-context adaptation close the gap.}
Methods that adapt using the unlabeled target batch at inference time provide the strongest gains. In particular, adapting normalization statistics (AdaBN) and entropy-minimizing BN adaptation (TENT) markedly improve robustness. Meta-learned in-context approaches further improve performance: ARM-BN achieves the best accuracy (Table~\ref{tab:results}), approaching the in-domain baseline and thereby largely neutralizing mild batch effects. This supports our hypothesis that, in microscopy, batch effects can often be corrected by \emph{fast, context-conditioned} adaptation rather than retraining.

\subsection{Strong domain shift: transfer across sources}
\label{sec:exp2_strong_shift}
Next, we evaluate whether the above conclusions hold under \emph{strong} domain shift across laboratories, which might have different data acquisition protocols and microscopes. We consider the setting where the main model is trained using labeled images obtained from only one source, i.e. a pharmaceutical company, and the evaluation is performed on samples from another source. Specifically, we use source 3 (S3) and source 8 (S8), and we perform experiments where the source (labeled) domain is S3 and the target domain is S8, and vice versa, and we also compare to the in-domain performance for each of them. Table~\ref{tab:generalization_full} reports MoA accuracy for each method under source transfer.

\textbf{Cross-source shifts are substantially harder.}
All methods degrade under cross-source transfer compared to within-source evaluation, confirming that laboratory-to-laboratory changes introduce stronger shifts than batch-to-batch drift within a single source.

\textbf{Normalization-based adaptation only at test time remains competitive but is less stable.}
AdaBN and TENT improve over non-adaptive baselines in some transfer directions, but the improvements are not consistent across all source$\rightarrow$target pairs. This variability is expected: under strong shift, the target feature statistics may differ not only due to style or illumination differences, but also due to more complex, structured differences (e.g., staining intensity distributions, microscope optics, and site-specific pipelines), and may require more robust context construction.

\textbf{Meta-learned in-context adaptation improves cross-source robustness.}
ARM-BN tends to provide stronger cross-source performance than purely test-time methods because the model is trained to adapt to new domains via episodic context sampling during training. Overall, Experiment~\ref{sec:exp2_strong_shift} highlights that cross-source robustness remains challenging, but \emph{learning to adapt} is a promising direction compared to static training or target-dependent retraining. 

\subsection{Realistic scenario 1: new experimental batch with label
shifts}
\label{sec:exp3_label_shift}
A known failure mode of BN-based adaptation is \emph{label shift} (i.e., changes in the class proportions between training and test) \citep{Park2023, Zhao2023}. In microscopy screening, this occurs naturally: many experiments are designed to test only a narrow set of hypotheses, and thus only a subset of MoAs may appear in a new batch. If the adaptation statistics are computed over the \emph{perturbed} samples, then changing the class mixture changes the feature statistics and can mislead adaptation.

\begin{table}[]
\centering
\caption{Comparison of MoA classification accuracy of different methods in a new experimental batch under different degrees of label shift. A smaller $\alpha$ value corresponds to a larger degree of label shift.}
\label{tab:label_shift}
\resizebox{\linewidth}{!}{
\begin{tabular}{lccc}
\toprule
\multirow{2}{*}{\textbf{Method}} & \multicolumn{3}{c}{\textbf{Label shift degree}}     \\
                          & $\alpha=1$ & $\alpha=0.1$ & $\alpha=0.01$ \\
\midrule
ERM (no adaptation)     & 0.862 ± 0.060 & 0.862 ± 0.061  & 0.862 ± 0.061 \\
\midrule
TENT                    & 0.832 ± 0.014 & 0.778 ± 0.016 & 0.156 ± 0.010 \\
DEYO                    & 0.830 ± 0.014 & 0.773 ± 0.015 & 0.156 ± 0.010 \\
ROID                    & 0.837 ± 0.017 & \textit{0.792} ± 0.020 & 0.157 ± 0.011 \\
AdaBN                  & \textit{0.851 ± 0.016}  & \textit{0.452 ± 0.010} & \textit{0.229 ± 0.009} \\
ARM-BN                  & 0.831 ± 0.017 & 0.416 ± 0.020 & 0.228 ± 0.009 \\
CS-ARM-BN (ours)        & \textbf{0.924 ± 0.018} & \textbf{0.906 ± 0.021} & \textbf{0.894 ± 0.022} \\
\bottomrule
\end{tabular}
}
\end{table}

\begin{table*}[]
\centering
\caption{Comparison of MoA classification accuracy of different methods across target sources (S8, S3, and S6) under different degrees of label shift. A lower $\alpha$ denotes a higher degree of label shift. \textbf{The accuracy values denote mean and standard deviation over re-runs with three different seeds.}}
\vspace{0.2em}
\label{tab:label_shift_new_source}
\begin{threeparttable}
\resizebox{\textwidth}{!}{%
\begin{tabular}{l cc cc cc cc}
\toprule
\multirow{2}{*}{\textbf{Method}} &
\multicolumn{2}{c}{\textbf{S3 $\rightarrow$ S8}} &
\multicolumn{2}{c}{\textbf{S8 $\rightarrow$ S3}} &
\multicolumn{2}{c}{\textbf{S3 $\rightarrow$ S6}} &
\multicolumn{2}{c}{\textbf{S8 $\rightarrow$ S6}} \\
\cmidrule(lr){2-3} \cmidrule(lr){4-5}
\cmidrule(lr){6-7} \cmidrule(lr){8-9}
 & $\alpha$ = 1 & $\alpha$ = 0.01  
 & $\alpha$ = 1 & $\alpha$ = 0.01
 & $\alpha$ = 1 & $\alpha$ = 0.01
 & $\alpha$ = 1 & $\alpha$ = 0.01 \\
\midrule

 no adaptation (BN) 
& 0.669±0.016 & 0.670±0.016
& 0.500±0.012 & 0.498±0.012
& 0.537±0.069 & 0.536±0.069
& 0.568±0.005 & 0.568±0.006 \\

\midrule
 TENT 
& 0.706±0.055 & 0.139±0.006 
& 0.659±0.005 & 0.136±0.000
& 0.552±0.027 & 0.135±0.004
& 0.571±0.006 & 0.122±0.001 \\

 DEYO 
& 0.705±0.056 & 0.139±0.006 
& 0.658±0.007 & 0.137±0.000
& 0.552±0.027 & 0.135±0.004
& 0.572±0.004 & 0.122±0.001 \\

 ROID 
& 0.715±0.050 & 0.138±0.004
& 0.663±0.005 & 0.135±0.000
& 0.557±0.025 & 0.135±0.004
& 0.574±0.006 & 0.123±0.001 \\

 AdaBN 
& 0.736±0.046 & 0.171±0.009 
& \textit{0.688±0.004} & 0.164±0.002
& \textit{0.580±0.023} & 0.159±0.006
& \textit{0.586±0.005} & 0.149±0.002 \\

BEN (w/o meta-learning)
& 0.125±0.000 & 0.125±0.000 
& 0.125±0.000 & 0.125±0.000 
& 0.124±0.001 & 0.122±0.004
& 0.125±0.000 & 0.125±0.000 \\

 ARM-BEN
& 0.664±0.031 & \textit{0.663±0.031}
& 0.544±0.031 & \textit{0.543±0.031}
& 0.547±0.040 &  \textit{0.546±0.040}
& 0.566±0.024 &  \textit{0.565±0.024} \\

 ARM-BN  
&\textit{ 0.747±0.010 }& 0.173±0.004
& 0.680±0.007 & 0.159±0.001
& 0.550±0.011 & 0.158±0.003
& 0.579±0.010 & 0.154±0.003 \\

 CS-ARM-BN (ours) 
& \textbf{0.878±0.006} & \textbf{0.825±0.007} 
& \textbf{0.788±0.014} & \textbf{0.730±0.015}
& \textbf{0.657±0.003} & \textbf{0.606±0.006}
& \textbf{0.648±0.011} & \textbf{0.611±0.015} \\

\bottomrule
\end{tabular}
}
\end{threeparttable}
\end{table*}
\textbf{Protocol.}
We simulate increasing levels of label shift by subsampling the target batch to enforce different class-mixture skew strengths. We report performance at three shift degrees (Table~\ref{tab:label_shift}): from mild to severe label shift. In all of the experiments, we use a constant number of labeled samples so the decrease in accuracy is not due to a smaller batch size. In every TTA method, the adaptation is unlabeled; only the class mixture is manipulated. While for TENT, AdaBN and ARM-BN, the accuracy decreases rapidly when label shift increases, the performance for CS-ARM-BN remains robust. Moreover, in Supplementary Table~\ref{apptab:label_shift_ablation} we also compare to AdaBN and ARM-BN including negative controls so that every method uses the same amount of samples to compute the BN statistics, and it can be seen that CS-ARM-BN still presents the highest performance.

\textbf{BN-adaptation degrades under label shift.}
As label shift increases, AdaBN, TENT and ARM-BN lose accuracy because their estimated BN statistics entangle (i) batch-specific nuisance variation with (ii) class-mixture-specific feature shifts. In other words, the adaptation signal becomes confounded by the changing MoA composition.

\textbf{Controls stabilize adaptation under label shift.}
CS-ARM-BN addresses this confounding by computing the adaptation statistics from both perturbed samples and \emph{negative controls}, which are independent of the MoA mixture by design. Consequently, the adaptation remains anchored to a consistent reference distribution per batch, and performance becomes substantially more stable across label-shift degrees. In Appendix Section \ref{appsec:exp3_batch_size}, we show that CS-ARM-BN stabilizes the performance under small batch sizes.

\subsection{Realistic scenario 2: new imaging source with label
shifts}
\label{sec:exp4_label_shift_new_source}
As in the previous experiment, the evaluation samples in this scenario exhibit a class distribution that differs from the training data. However, in this case, the domain shift is more pronounced, as the test samples originate from a new imaging source. In Table~\ref{tab:label_shift_new_source}, we show that, similarly to Experiment~\ref{sec:exp3_label_shift}, the performance of methods that do not leverage negative controls (TENT, AdaBN, and ARM-BN) degrades as the degree of label shift increases.

We evaluate the performance of BEN, which relies exclusively on control samples for adaptation, for which we also extend BEN to meta-learning. While the version without meta-learning performs poorly, the meta-learned variant, despite achieving lower accuracy than CS-ARM-BN, outperforms the other methods under strong label shift.

Consistent with the previous experiment, CS-ARM-BN stabilizes the batch normalization statistics and achieves the highest accuracy across both mild and severe domain shifts.

\section{Discussion and Limitations}
We framed batch effect correction for mechanism of action (MoA) classification as a \emph{multi-source domain adaptation} problem, reflecting a real use-case scenario of high throughput microscopy imaging data. Our study shows that \emph{in-context adaptation}, where models condition their parameters on a small set of unlabeled samples from a new batch, outperforms established approaches.

\paragraph{Limitations.} Although we have shown that CS-ARM-BN presents a robust performance, especially under label shifts and when the number of perturbed samples is small, our study is limited to the experimental settings considered here and does not cover other scenarios. For example, we did not consider the setting where the controls come from a different cell line than the perturbed samples. Another noteworthy experimental setting is that we process each image from a certain well individually. Processing all well images together may make the results more stable, alleviating the impact of noisy views. Moreover, additional strategies like test-time scaling might further improve performance, as shown in Supplementary Section~\ref{appsec:test_time_scaling}.

\paragraph{Experimental findings.} Across four experiments, we find that (i) batch effects cause a large generalization gap even under mild within-source shifts, (ii) cross-source transfer is substantially harder but benefits from meta-learned in-context adaptation, and (iii) standard BN-based test-time adaptation methods are weak when target statistics are estimated from few or label-skewed perturbed samples. By leveraging the experimental design of high-content screening, i.e. the presence of negative controls in every batch, CS-ARM-BN yields strong accuracy while remaining robust to small context sizes and label shift. We intentionally focus on lightweight, fast adaptation methods that can operate with limited target data and minimal computation, reflecting realistic deployment conditions. 

In the evaluated settings, ARM-BN and CS-ARM-BN nearly close the performance gap between in-domain and out-of-domain evaluation. On unseen experimental batches, ARM-BN and CS-ARM-BN reach $0.935 \pm 0.018$ and $0.930 \pm 0.019$ accuracy, respectively, matching the $0.939 \pm 0.005$ accuracy observed within the training distribution. This demonstrates that batch effects, long a bottleneck for robust MoA prediction, can be effectively neutralized. Our findings highlight that in-context test-time domain adaptation methods can make deep learning models for MoA identification not only accurate but also practical for deployment across experimental batches.

\section*{Software and data}
Code and data are available at \url{https://github.com/ml-jku/cs-arm-bn}.

\section*{Acknowledgments}
The ELLIS Unit Linz, the LIT AI Lab, the Institute for
Machine Learning, are supported by the Federal State Up-
per Austria. We thank the projects FWF AIRI FG 9-N
(10.55776/FG9), AI4GreenHeatingGrids (FFG- 899943),
Stars4Waters (HORIZON-CL6-2021-CLIMATE-01-01),
FWF Bilateral Artificial Intelligence (10.55776/COE12).
We thank Merck Healthcare KGaA, NXAI GmbH, Audi
AG, TÜV Holding GmbH, Software Competence Center Hagenberg GmbH, dSPACE GmbH, TRUMPF SE + Co. KG.
TP is funded by the European Union, EIC-2023-PATHFINDEROPEN-01 (I-SCREEN, grant no.
101130093).
Views and opinions expressed are however those of the author(s) only and do not necessarily reflect those of the European Union or European Innovation Council and SMEs Executive Agency (EISMEA). Neither the European Union nor the granting authority can be held responsible for them.

\section*{Impact Statement}
This work addresses one of the central barriers to deploying machine learning in real-world biomedical settings: poor generalization across experimental batches and domains. We show that widely used domain adaptation strategies can fail under realistic conditions involving batch effects and label shift, and we introduce CS-ARM-BN, a method that explicitly leverages negative control samples to estimate and correct domain-specific shifts in a principled and stable way. 

The impact of this work is both methodological and translational. Methodologically, it highlights the limitations of assuming covariate shift alone in scientific imaging data and demonstrates how experimental design elements (negative controls) can be incorporated directly into training and inference pipelines to adress these limitations. Translationally, improved cross-batch generalization enables more reliable mechanism-of-action classification. Also, by reducing the need for extensive per-batch retraining and manual correction, our approach could lower the cost of model deployment in comparison to other domain adaptation methods. 

However, in case of failure of any machine learning model employed for drug discovery, false positives can lead to costly downstream validation of ineffective compounds, while false negatives risk discarding potentially valuable therapeutic candidates early in the pipeline. Moreover, our evaluation focuses on a limited set of compounds with well-characterized and relatively strong mechanisms of action, where phenotypic signals are pronounced. Compounds with subtler effects might be more difficult to classify. 

Despite these limitations, this work contributes to the development of machine learning systems that are robust to non-i.i.d. data, which is a requirement for trustworthy AI in biology, medicine, and other experimental sciences.
\bibliographystyle{apalike}
\bibliography{refs}

\begin{thebibliography}{}

\bibitem[Ando et~al., 2017]{Ando2017}
Ando, D.~M., McLean, C.~Y., and Berndl, M. (2017).
\newblock Improving phenotypic measurements in high-content imaging screens.
\newblock {\em bioRxiv}, page 161422.

\bibitem[Arevalo et~al., 2024]{Arevalo2024}
Arevalo, J., Su, E., Ewald, J.~D., van Dijk, R., Carpenter, A.~E., and Singh, S. (2024).
\newblock Evaluating batch correction methods for image-based cell profiling.
\newblock {\em Nature Communications 2024 15:1}, 15:1--12.

\bibitem[Baxter, 1998]{baxter1998theoretical}
Baxter, J. (1998).
\newblock Theoretical models of learning to learn.
\newblock In {\em Learning to learn}, pages 71--94. Springer.

\bibitem[Ben-David et~al., 2006]{ben2006analysis}
Ben-David, S., Blitzer, J., Crammer, K., and Pereira, F. (2006).
\newblock Analysis of representations for domain adaptation.
\newblock {\em Advances in neural information processing systems}, 19.

\bibitem[Blanchard et~al., 2011]{NIPS2011_b571ecea}
Blanchard, G., Lee, G., and Scott, C. (2011).
\newblock Generalizing from several related classification tasks to a new unlabeled sample.
\newblock In Shawe-Taylor, J., Zemel, R., Bartlett, P., Pereira, F., and Weinberger, K., editors, {\em Advances in Neural Information Processing Systems}, volume~24. Curran Associates, Inc.

\bibitem[Boudiaf et~al., 2022]{Boudiaf2022}
Boudiaf, M., Mueller, R., Ayed, I.~B., and Bertinetto, L. (2022).
\newblock Parameter-free online test-time adaptation.
\newblock {\em Proceedings of the IEEE Computer Society Conference on Computer Vision and Pattern Recognition}, 2022-June:8334--8343.

\bibitem[Bousmalis et~al., 2017]{Bousmalis2017}
Bousmalis, K., Silberman, N., Dohan, D., Erhan, D., and Krishnan, D. (2017).
\newblock Unsupervised pixel-level domain adaptation with generative adversarial networks.
\newblock {\em Proceedings - 30th IEEE Conference on Computer Vision and Pattern Recognition, CVPR 2017}, 2017-January:95--104.

\bibitem[Bray et~al., 2017]{Bray2017}
Bray, M.-A., Carpenter, A., of~MIT, B.~I., and Platform, H.~I. (2017).
\newblock Advanced assay development guidelines for image-based high content screening and analysis.
\newblock {\em Assay Guidance Manual}.

\bibitem[{Carpenter-Singh and Cimini Labs at the Broad Institute}, 2026]{CPgallery}
{Carpenter-Singh and Cimini Labs at the Broad Institute} (2026).
\newblock \emph{CellPainting {G}allery}.
\newblock \url{https://registry.opendata.aws/cellpainting-gallery/}.
\newblock [Online; accessed 26-May-2026].

\bibitem[Chandrasekaran et~al., 2023]{Chandrasekaran2023}
Chandrasekaran, S.~N., Ackerman, J., Alix, E., Ando, D.~M., Arevalo, J., Bennion, M., Boisseau, N., Borowa, A., Boyd, J.~D., Brino, L., Byrne, P.~J., Ceulemans, H., Ch’ng, C., Cimini, B.~A., Clevert, D.-A., Deflaux, N., Doench, J.~G., Dorval, T., Doyonnas, R., Dragone, V., Engkvist, O., Faloon, P.~W., Fritchman, B., Fuchs, F., Garg, S., Gilbert, T.~J., Glazer, D., Gnutt, D., Goodale, A., Grignard, J., Guenther, J., Han, Y., Hanifehlou, Z., Hariharan, S., Hernandez, D., Horman, S.~R., Hormel, G., Huntley, M., Icke, I., Iida, M., Jacob, C.~B., Jaensch, S., Khetan, J., Kost-Alimova, M., Krawiec, T., Kuhn, D., Lardeau, C.-H., Lembke, A., Lin, F., Little, K.~D., Lofstrom, K.~R., Lotfi, S., Logan, D.~J., Luo, Y., Madoux, F., Zapata, P. A.~M., Marion, B.~A., Martin, G., McCarthy, N.~J., Mervin, L., Miller, L., Mohamed, H., Monteverde, T., Mouchet, E., Nicke, B., Ogier, A., Ong, A.-L., Osterland, M., Otrocka, M., Peeters, P.~J., Pilling, J., Prechtl, S., Qian, C., Rataj, K., Root, D.~E., Sakata, S.~K., Scrace, S.,
  Shimizu, H., Simon, D., Sommer, P., Spruiell, C., Sumia, I., Swalley, S.~E., Terauchi, H., Thibaudeau, A., Unruh, A., de~Waeter, J.~V., Dyck, M.~V., van Staden, C., Warchoł, M., Weisbart, E., Weiss, A., Wiest-Daessle, N., Williams, G., Yu, S., Zapiec, B., Żyła, M., Singh, S., and Carpenter, A.~E. (2023).
\newblock Jump cell painting dataset: morphological impact of 136,000 chemical and genetic perturbations.
\newblock {\em bioRxiv}, page 2023.03.23.534023.

\bibitem[Chen et~al., 2020]{Chen2020}
Chen, W., Zhao, Y., Chen, X., Yang, Z., Xu, X., Bi, Y., Chen, V., Li, J., Choi, H., Ernest, B., Tran, B., Mehta, M., Kumar, P., Farmer, A., Mir, A., Mehra, U.~A., Li, J.~L., Moos, M., Xiao, W., and Wang, C. (2020).
\newblock A multicenter study benchmarking single-cell rna sequencing technologies using reference samples.
\newblock {\em Nature Biotechnology 2020 39:9}, 39:1103--1114.

\bibitem[Chung et~al., 2024]{Chung2024}
Chung, J., Hyun, S., and Heo, J.~P. (2024).
\newblock Style injection in diffusion: A training-free approach for adapting large-scale diffusion models for style transfer.
\newblock {\em Proceedings of the IEEE Computer Society Conference on Computer Vision and Pattern Recognition}, pages 8795--8805.

\bibitem[Dong et~al., 2026]{Dong2026.01.09.698608}
Dong, M., Adduri, A., Gautam, D., Carpenter, C., Shah, R., Ricci-Tam, C., Kluger, Y., Burke, D.~P., and Roohani, Y.~H. (2026).
\newblock Stack: In-context learning of single-cell biology.
\newblock {\em bioRxiv}.

\bibitem[Farahani et~al., 2020]{Farahani2020}
Farahani, A., Voghoei, S., Rasheed, K., and Arabnia, H.~R. (2020).
\newblock A brief review of domain adaptation.
\newblock {\em ArXiv}, pages 877--894.

\bibitem[Finn et~al., 2017]{finn2017model}
Finn, C., Abbeel, P., and Levine, S. (2017).
\newblock Model-agnostic meta-learning for fast adaptation of deep networks.
\newblock In {\em International conference on machine learning}, pages 1126--1135. PMLR.

\bibitem[Ganin and Lempitsky, 2014]{ganin2014}
Ganin, Y. and Lempitsky, V. (2014).
\newblock Unsupervised domain adaptation by backpropagation.
\newblock {\em 32nd International Conference on Machine Learning, ICML 2015}, 2:1180--1189.

\bibitem[Gong et~al., 2022]{Gong2022}
Gong, T., Jeong, J., Kim, T., Kim, Y., Shin, J., and Lee, S.~J. (2022).
\newblock Note: Robust continual test-time adaptation against temporal correlation.
\newblock {\em Advances in Neural Information Processing Systems}, 35.

\bibitem[Graham et~al., 2021]{Graham2021}
Graham, B., El-Nouby, A., Touvron, H., Stock, P., Joulin, A., Jégou, H., and Douze, M. (2021).
\newblock Levit: a vision transformer in convnet's clothing for faster inference.
\newblock {\em Proceedings of the IEEE International Conference on Computer Vision}, pages 12239--12249.

\bibitem[Guo et~al., 2018]{guo2018multi}
Guo, J., Shah, D.~J., and Barzilay, R. (2018).
\newblock Multi-source domain adaptation with mixture of experts.
\newblock {\em arXiv preprint arXiv:1809.02256}.

\bibitem[Haghverdi et~al., 2018]{Haghverdi2018}
Haghverdi, L., Lun, A.~T., Morgan, M.~D., and Marioni, J.~C. (2018).
\newblock Batch effects in single-cell rna-sequencing data are corrected by matching mutual nearest neighbors.
\newblock {\em Nature Biotechnology 2018 36:5}, 36:421--427.

\bibitem[Haslum et~al., 2023]{Haslum2023}
Haslum, J.~F., Matsoukas, C., Leuchowius, K.~J., and Smith, K. (2023).
\newblock Bridging generalization gaps in high content imaging through online self-supervised domain adaptation.
\newblock {\em IEEE Workshop/Winter Conference on Applications of Computer Vision}, pages 7723--7732.

\bibitem[He et~al., 2015]{He2015}
He, K., Zhang, X., Ren, S., and Sun, J. (2015).
\newblock Deep residual learning for image recognition.
\newblock {\em Proceedings of the IEEE Computer Society Conference on Computer Vision and Pattern Recognition}, 2016-December:770--778.

\bibitem[Hie et~al., 2019]{Hie2019}
Hie, B., Bryson, B., and Berger, B. (2019).
\newblock Efficient integration of heterogeneous single-cell transcriptomes using scanorama.
\newblock {\em Nature biotechnology}, 37:685--691.

\bibitem[Hochreiter et~al., 2001]{hochreiter2001learning}
Hochreiter, S., Younger, A.~S., and Conwell, P.~R. (2001).
\newblock Learning to learn using gradient descent.
\newblock In {\em International conference on artificial neural networks}, pages 87--94. Springer.

\bibitem[Hughes et~al., 2011]{Hughes2011}
Hughes, J.~P., Rees, S.~S., Kalindjian, S.~B., and Philpott, K.~L. (2011).
\newblock Principles of early drug discovery.
\newblock {\em British Journal of Pharmacology}, 162:1239.

\bibitem[Ioffe and Szegedy, 2015]{Sergei2015}
Ioffe, S. and Szegedy, C. (2015).
\newblock Batch normalization: accelerating deep network training by reducing internal covariate shift.
\newblock In {\em Proceedings of the 32nd International Conference on International Conference on Machine Learning - Volume 37}, ICML'15, page 448–456. JMLR.org.

\bibitem[Johnson et~al., 2007]{Johnson2007}
Johnson, W.~E., Li, C., and Rabinovic, A. (2007).
\newblock Adjusting batch effects in microarray expression data using empirical bayes methods.
\newblock {\em Biostatistics (Oxford, England)}, 8:118--127.

\bibitem[Kim et~al., 2025]{Kim2025}
Kim, V., Adaloglou, N., Osterland, M., Morelli, F.~M., Halawa, M., König, T., Gnutt, D., and Zapata, P. A.~M. (2025).
\newblock Self-supervision advances morphological profiling by unlocking powerful image representations.
\newblock {\em Scientific Reports 2025 15:1}, 15:1--15.

\bibitem[Kimura and Hino, 2024]{kimura2024short}
Kimura, M. and Hino, H. (2024).
\newblock A short survey on importance weighting for machine learning.
\newblock {\em arXiv preprint arXiv:2403.10175}.

\bibitem[Knowles and Gromo, 2003]{Knowles2003}
Knowles, J. and Gromo, G. (2003).
\newblock A guide to drug discovery: Target selection in drug discovery.
\newblock {\em Nature reviews. Drug discovery}, 2:63--69.

\bibitem[Korsunsky et~al., 2019]{Korsunsky2019}
Korsunsky, I., Millard, N., Fan, J., Slowikowski, K., Zhang, F., Wei, K., Baglaenko, Y., Brenner, M., Loh, P., and Raychaudhuri, S. (2019).
\newblock Fast, sensitive and accurate integration of single-cell data with harmony.
\newblock {\em Nature Methods 2019 16:12}, 16:1289--1296.

\bibitem[Kraus et~al., 2024]{Kraus2024}
Kraus, O., Kenyon-Dean, K., Saberian, S., Fallah, M., McLean, P., Leung, J., Sharma, V., Khan, A., Balakrishnan, J., Celik, S., Beaini, D., Sypetkowski, M., Cheng, C.~V., Morse, K., Makes, M., Mabey, B., and Earnshaw, B. (2024).
\newblock Masked autoencoders for microscopy are scalable learners of cellular biology.
\newblock {\em Proceedings of the IEEE Computer Society Conference on Computer Vision and Pattern Recognition}, pages 11757--11768.

\bibitem[Lee et~al., 2024]{Lee2024}
Lee, J., Jung, D., Lee, S., Park, J., Shin, J., Hwang, U., and Yoon, S. (2024).
\newblock Entropy is not enough for test-time adaptation: From the perspective of disentangled factors.
\newblock {\em 12th International Conference on Learning Representations, ICLR 2024}.

\bibitem[Leek et~al., 2010]{Leek2010}
Leek, J.~T., Scharpf, R.~B., Bravo, H.~C., Simcha, D., Langmead, B., Johnson, W.~E., Geman, D., Baggerly, K., and Irizarry, R.~A. (2010).
\newblock Tackling the widespread and critical impact of batch effects in high-throughput data.
\newblock {\em Nature reviews. Genetics}, 11:10.1038/nrg2825.

\bibitem[Li et~al., 2016]{Li2016}
Li, Y., Wang, N., Shi, J., Liu, J., and Hou, X. (2016).
\newblock Revisiting batch normalization for practical domain adaptation.
\newblock {\em International Conference on Learning Representations}.

\bibitem[Lin and Lu, 2022]{Lin2022}
Lin, A. and Lu, A. (2022).
\newblock Incorporating knowledge of plates in batch normalization improves generalization of deep learning for microscopy images.
\newblock In Knowles, D.~A., Mostafavi, S., and Lee, S.-I., editors, {\em Proceedings of the 17th Machine Learning in Computational Biology meeting}, volume 200 of {\em Proceedings of Machine Learning Research}, pages 74--93. PMLR.

\bibitem[Liu and Tuzel, 2016]{NIPS2016_502e4a16}
Liu, M.-Y. and Tuzel, O. (2016).
\newblock Coupled generative adversarial networks.
\newblock In Lee, D., Sugiyama, M., Luxburg, U., Guyon, I., and Garnett, R., editors, {\em Advances in Neural Information Processing Systems}, volume~29. Curran Associates, Inc.

\bibitem[Lopez et~al., 2018]{Lopez2018}
Lopez, R., Regier, J., Cole, M.~B., Jordan, M.~I., and Yosef, N. (2018).
\newblock Deep generative modeling for single-cell transcriptomics.
\newblock {\em Nature methods}, 15:1053--1058.

\bibitem[Luecken et~al., 2021]{Luecken2021}
Luecken, M.~D., Büttner, M., Chaichoompu, K., Danese, A., Interlandi, M., Mueller, M.~F., Strobl, D.~C., Zappia, L., Dugas, M., Colomé-Tatché, M., and Theis, F.~J. (2021).
\newblock Benchmarking atlas-level data integration in single-cell genomics.
\newblock {\em Nature Methods 2021 19:1}, 19:41--50.

\bibitem[Marsden et~al., 2023]{Marsden2023}
Marsden, R.~A., Döbler, M., and Yang, B. (2023).
\newblock Universal test-time adaptation through weight ensembling, diversity weighting, and prior correction.
\newblock {\em Proceedings - 2024 IEEE Winter Conference on Applications of Computer Vision, WACV 2024}, pages 2543--2553.

\bibitem[Palma et~al., 2025]{Palma2025}
Palma, A., Theis, F.~J., and Lotfollahi, M. (2025).
\newblock Predicting cell morphological responses to perturbations using generative modeling.
\newblock {\em Nature Communications}, 16:1--19.

\bibitem[Park et~al., 2023]{Park2023}
Park, S., Yang, S., Choo, J., and Yun, S. (2023).
\newblock Label shift adapter for test-time adaptation under covariate and label shifts.
\newblock {\em Proceedings of the IEEE International Conference on Computer Vision}, pages 16375--16385.

\bibitem[Peidli et~al., 2023]{Peidli2023}
Peidli, S., Green, T.~D., Shen, C., Gross, T., Min, J., Garda, S., Yuan, B., Schumacher, L.~J., Taylor-King, J.~P., Marks, D.~S., Luna, A., Blüthgen, N., and Sander, C. (2023).
\newblock sc{P}erturb: Harmonized single-cell perturbation data.
\newblock {\em bioRxiv}, page 2022.08.20.504663.

\bibitem[Peng et~al., 2019]{peng2019moment}
Peng, X., Bai, Q., Xia, X., Huang, Z., Saenko, K., and Wang, B. (2019).
\newblock Moment matching for multi-source domain adaptation.
\newblock In {\em Proceedings of the IEEE/CVF international conference on computer vision}, pages 1406--1415.

\bibitem[Shimodaira, 2000]{shimodaira2000improving}
Shimodaira, H. (2000).
\newblock Improving predictive inference under covariate shift by weighting the log-likelihood function.
\newblock {\em Journal of statistical planning and inference}, 90(2):227--244.

\bibitem[Stirling et~al., 2021]{Stirling2021}
Stirling, D.~R., Swain-Bowden, M.~J., Lucas, A.~M., Carpenter, A.~E., Cimini, B.~A., and Goodman, A. (2021).
\newblock Cellprofiler 4: improvements in speed, utility and usability.
\newblock {\em BMC Bioinformatics}, 22:433.

\bibitem[Stuart et~al., 2019]{Stuart2019}
Stuart, T., Butler, A., Hoffman, P., Hafemeister, C., Papalexi, E., Mauck, W.~M., Hao, Y., Stoeckius, M., Smibert, P., and Satija, R. (2019).
\newblock Comprehensive integration of single-cell data.
\newblock {\em Cell}, 177:1888--1902.e21.

\bibitem[Sun et~al., 2015]{sun2015}
Sun, B., Feng, J., and Saenko, K. (2015).
\newblock Return of frustratingly easy domain adaptation.
\newblock {\em 30th AAAI Conference on Artificial Intelligence, AAAI 2016}, pages 2058--2065.

\bibitem[Sypetkowski et~al., 2023]{Sypetkowski2023}
Sypetkowski, M., Rezanejad, M., Saberian, S., Kraus, O., Urbanik, J., Taylor, J., Mabey, B., Victors, M., Yosinski, J., Sereshkeh, A.~R., Haque, I., and Earnshaw, B. (2023).
\newblock {R}x{R}x1: A dataset for evaluating experimental batch correction methods.
\newblock {\em IEEE Computer Society Conference on Computer Vision and Pattern Recognition Workshops}, 2023-June:4285--4294.

\bibitem[Taigman et~al., 2016]{Taigman2016UnsupervisedCI}
Taigman, Y., Polyak, A., and Wolf, L. (2016).
\newblock Unsupervised cross-domain image generation.
\newblock {\em ArXiv}, abs/1611.02200.

\bibitem[Venkat et~al., 2020]{venkat2020your}
Venkat, N., Kundu, J.~N., Singh, D., Revanur, A., et~al. (2020).
\newblock Your classifier can secretly suffice multi-source domain adaptation.
\newblock {\em Advances in Neural Information Processing Systems}, 33:4647--4659.

\bibitem[Wang et~al., 2021]{wang2021tent}
Wang, D., Shelhamer, E., Liu, S., Olshausen, B., and Darrell, T. (2021).
\newblock {TENT}: Fully test-time adaptation by entropy minimization.
\newblock In {\em International Conference on Learning Representations}.

\bibitem[Wang et~al., 2022]{Wang2022}
Wang, Q., Fink, O., Gool, L.~V., and Dai, D. (2022).
\newblock Continual test-time domain adaptation.
\newblock {\em Proceedings of the IEEE Computer Society Conference on Computer Vision and Pattern Recognition}, 2022-June:7191--7201.

\bibitem[Wen et~al., 2020]{wen2020domain}
Wen, J., Greiner, R., and Schuurmans, D. (2020).
\newblock Domain aggregation networks for multi-source domain adaptation.
\newblock In {\em International conference on machine learning}, pages 10214--10224. PMLR.

\bibitem[Yang et~al., 2020]{yang2020curriculum}
Yang, L., Balaji, Y., Lim, S.-N., and Shrivastava, A. (2020).
\newblock Curriculum manager for source selection in multi-source domain adaptation.
\newblock In {\em European conference on computer vision}, pages 608--624. Springer.

\bibitem[Zellinger et~al., 2017]{Zellinger2017cmd}
Zellinger, W., Grubinger, T., Lughofer, E., Natschl{\"a}ger, T., and Saminger-Platz, S. (2017).
\newblock Central moment discrepancy (cmd) for domain-invariant representation learning.
\newblock In {\em International Conference on Learning Representations (ICLR)}.

\bibitem[Zhang et~al., 2025]{Zhang2025.02.20.639398}
Zhang, J., Ubas, A.~A., de~Borja, R., Svensson, V., Thomas, N., Thakar, N., Lai, I., Winters, A., Khan, U., Jones, M.~G., Tran, V., Pangallo, J., Papalexi, E., Sapre, A., Nguyen, H., Sanderson, O., Nigos, M., Kaplan, O., Schroeder, S., Hariadi, B., Marrujo, S., Salvino, C. C.~A., Gallareta~Olivares, G., Koehler, R., Geiss, G., Rosenberg, A., Roco, C., Merico, D., Alidoust, N., Goodarzi, H., and Yu, J. (2025).
\newblock {T}ahoe-100{M}: A giga-scale single-cell perturbation atlas for context-dependent gene function and cellular modeling.
\newblock {\em bioRxiv}.

\bibitem[Zhang et~al., 2021]{Zhang2021}
Zhang, M., Marklund, H., Dhawan, N., Gupta, A., Levine, S., and Finn, C. (2021).
\newblock Adaptive {R}isk {M}inimization: learning to adapt to domain shift.
\newblock In {\em Proceedings of the 35th International Conference on Neural Information Processing Systems}, NeurIPS '21, Red Hook, NY, USA. Curran Associates Inc.

\bibitem[Zhao et~al., 2023]{Zhao2023}
Zhao, H., Liu, Y., Alahi, A., and Lin, T. (2023).
\newblock On pitfalls of test-time adaptation.
\newblock {\em Proceedings of Machine Learning Research}, 202:42058--42080.

\bibitem[Zhao et~al., 2018]{zhao2018adversarial}
Zhao, H., Zhang, S., Wu, G., Moura, J.~M., Costeira, J.~P., and Gordon, G.~J. (2018).
\newblock Adversarial {M}ultiple {S}ource {D}omain {A}daptation.
\newblock {\em Advances in neural information processing systems}, 31.

\end{thebibliography}

\appendix
\onecolumn
\resumetocwriting

\renewcommand{\topfraction}{0.95}
\renewcommand{\bottomfraction}{0.95}
\renewcommand{\textfraction}{0.05}
\renewcommand{\floatpagefraction}{0.85}
\tableofcontents

\newpage

\renewcommand{\thetable}{{A\arabic{table}}}
\renewcommand{\thefigure}{{A\arabic{figure}}}

\section{Notation table}
\begin{table}[H]
\resizebox{\textwidth}{!}{%
\begin{tabular}{l c l}
\toprule
\textbf{Definition} & \textbf{Symbol} & \textbf{Type} \\
\midrule
\multicolumn{3}{l}{\textbf{Scalars}} \\
number of source domains (batches) & $B$ & $\mathbb{N}$ \\
number of samples per source domain (simplified) & $N$ & $\mathbb{N}$ \\
number of samples from one of the target domains & $M$ & $\mathbb{N}$ \\
number of control samples from target domain & $C$ & $\mathbb{N}$ \\
number of non-control samples from target domain & $L$ & $\mathbb{N}$ \\
feature dimension & $D$ & $\mathbb{N}$ \\
number of classes & $K$ & $\mathbb{N}$ \\
DANN domain-loss trade-off & $\lambda$ & $\mathbb{R}_{\ge 0}$ \\
CORAL penalty weight & $\gamma$ & $\mathbb{R}_{\ge 0}$ \\
\midrule
\multicolumn{3}{l}{\textbf{Sets, spaces, and distributions}} \\
input/image space & $\mathcal X$ &  \\
label space / class set & $\mathcal Y$ &  \\
training data (all sources) & $\mathcal D^{\mathrm{train}}$ & $\bigcup_{b=1}^{B}\mathcal D^{(b)}$ \\
data from source domain $b$ & $\mathcal D^{(b)}$ & $\{(\Bx^{(b)}_n,\By^{(b)}_n)\}_{n=1}^{N}$ \\
unseen target-batch data & $\mathcal D^{\beta}$ & $\{\Bx^{\beta}_1,\dots,\Bx^{\beta}_M\}$ \\
domain-specific data distribution & $p^{b}_{xy}$ & distribution on $\mathcal X\times\mathcal Y$ \\
meta-distribution over domains & $\mu$ & distribution on domains \\
\midrule
\multicolumn{3}{l}{\textbf{Representations and tensors}} \\
image / sample / input & $\Bx$ & $\mathcal X$ \\
control image / control sample & $\Bz$ & $\mathcal X$ \\
label & $\By$ & $\mathcal Y$ \\
feature embedding & $\Ba=\Bf(\Bx;\Bw')$ & $\mathbb{R}^{D}$ \\
TVN-normalized embedding & $\tilde{\Ba}$ & $\mathbb{R}^{D}$ \\
control embeddings (batch $\beta$) & $\{\Bz^{\beta}_j\}$ &  $\mathbb{R}^{D}$ \\
source/target feature covariances & $\BC_S,\,\BC_T$ & $\mathbb{R}^{D\times D}$ \\
control mean/covariance (TVN) & $\Bmu^{\mathrm{ctrl},\beta},\,\BSigma^{\mathrm{ctrl},\beta}$ & $\mathbb{R}^{D},\,\mathbb{R}^{D\times D}$ \\
\midrule
\multicolumn{3}{l}{\textbf{Models and parameters}} \\
feature extractor & $\Bf$, $\Bf_{\Bw}$,$\Bf(\cdot;\Bw')$ & $\mathcal X\to\mathbb{R}^{D}$ \\
classifier & $\Bc(\cdot;\Btheta)$ & $\mathbb{R}^{D}\to \mathcal Y$ \\
prediction model & $\Bg(\Bx;\Bw)=(\Bc\circ\Bf)(\Bx;\Bw)$ & $\mathcal X\to \mathcal Y$ \\
model parameters & $\Bw$  & $\mathcal W$ \\
adaptation function (test-time) & $\Bh_{\Bphi}(\Bw,\Bx^{\beta}_{1:M})$ & $\mathcal W\times\mathcal X^{M}\to\mathcal W$ \\
adaptation parameters & $\Bphi$ & parameters of $\Bh$ \\
adapted parameters for batch $\beta$ & $\Bw^{\beta}$ & $\mathcal W$ \\
domain discriminator (DANN) & $\Bd_{\Bv}$ & $\mathbb{R}^{D}\to \mathcal S^{B-1}$ \\
\midrule
\multicolumn{3}{l}{\textbf{Losses and objectives}} \\
supervised loss (cross-entropy) & $\mathcal L_{\mathrm{CE}}(y,\Bg(\Bx;\Bw))$ & $\mathbb{R}$ \\
domain loss (DANN) & $\mathcal L_{\mathrm{CE}}(b,\Bd_{\Bv}(\Bh))$ & $\mathbb{R}$ \\
CORAL discrepancy & $\|\BC_S - \BC_T\|_F^2$ & $\mathbb{R}_{\ge 0}$ \\
\midrule
\multicolumn{3}{l}{\textbf{BatchNorm (BN) parameters}} \\
test-batch mean/var (per channel, batch $\beta$) & $\Bmu_{\mathrm{BN}}, \Bsigma_{\mathrm{BN}}$  & $\mathbb{R}^D,\,\mathbb{R}^D_{>0}$ \\
affine parameters & $\kappa, \lambda$  & $\mathbb{R}^D,\,\mathbb{R}^D$ \\
\bottomrule
\end{tabular}
}
\end{table}

\clearpage

\section{Training details}
In our experiments, we used either Nvidia A40 or A100 GPUs and all experiments use a single GPU. When reporting results, for the new experimental batch evaluation (experiment \ref{sec:exp1_mild_shift}) and for the in-domain evaluation of source 3 (S3$\rightarrow$S3 in Table \ref{tab:generalization_full}), we perform 5-fold cross-validation and report the mean and standard deviation across folds. For the in-domain evaluation of source 8 (S8$\rightarrow$S8), due to computational constrains, we perform re-runs in two different random splits. 

For all methods, in Experiments \ref{sec:exp1_mild_shift} and \ref{sec:exp3_label_shift} we set a maximum number of training epochs to 400 and implement an early stopping strategy, setting the patience to 80 epochs, so that if the validation loss does not decrease with respect to the current minimum loss for 80 epochs, the training is stopped. In Experiments \ref{sec:exp2_strong_shift} and \ref{sec:exp4_label_shift_new_source}, due to computational constraints, we set a maximum number of training epochs to 200 and a patience of 20. In all experiments, the checkpoint used for evaluating the test set is the one that achieved the lowest validation loss.  By default, we use a ResNet50 as a feature encoder. As an optimizer, unless otherwise stated, we use Adam and the selected learning rate is decreased using a cosine annealing scheduler. The hyperparameters were manually tuned, and the exploration space for each method can be found in Supplementary Table~\ref{tab:hyperparams}

For the methods based on the CA-MAE foundation model, linear probing is performed, so that the weights of the feature encoder are frozen and only the classifier, which consists in one linear layer, is trained. The pretrained weights of this model are the weights of a ViT-S/16, provided in \hyperlink{https://huggingface.co/recursionpharma/OpenPhenom}{HuggingFace} by \citet{Kraus2024}, trained with the RxRx3 dataset and subsets of the JUMP-CP with samples treated for gene knockout and overexpression.

For the entropy-based methods TENT \citep{wang2021tent}, DEYO \citep{Lee2024} and ROID \cite{Marsden2023}, we consider every plate as a new domain, encoding one full plate at a time and performing 3 update steps per plate, with a learning rate of 0.001. 

In the style transfer baseline, as the images were generated using a pre-trained diffusion model, the input for the feature extractor consisted in 3 channels instead of 5. Moreover, for computational efficiency, the images were generated with a resolution of 256x256. 

For performing the TVN correction, control images from the corresponding matching plate of each perturbed sample are used for centering and whitening.

\begin{table*}[htp]
    \centering
      \caption{Considered hyperparameter space for all compared methods.
      The selected configurations for mechanism of action prediction
      based on manual search on validation set are shown in bold.}
    \resizebox{\textwidth}{!}{
    \begin{tabular}{@{}llc@{}}
        \toprule
        & \textbf{Hyperparameter} & \textbf{Explored space}\\
        \midrule
        \multirow{3}{7em}{Baseline (ResNet)} & Learning rate & \{\textbf{0.001}, 0.005\} \\
                                    & Scheduler & \{\textbf{Cosine annealing}, None\} \\
                                    & Batch size  & \{64, \textbf{128}\} \\
        \midrule
        \multirow{2}{*}{Baseline (CA-MAE)} &  Learning rate & \{0.001, 0.002, 0.003, 0.004, 0.005, 0.006, 0.007, \textbf{0.009}, 0.01\}  \\
        & Batch size & \{128\}\ \\ 
        \midrule
        \multirow{4}{7em}{DANN}    & Learning rate & \{0.001\} \\

                                    & Batch size & \{128\} \\
                                    & $\lambda$ & \{0.1, 0.5, 0.7, \textbf{2}, 3, 4\} \\
                                    & Domain classifier learning rate & \{0.0001\} \\
                                    & Unlabeled batch size & \{128\} \\

        \midrule
        \multirow{4}{7em}{CORAL}    & Learning rate & \{0.001\} \\
                                    & Batch size & \{128\} \\
                                    & $\gamma$ & \{\textbf{0.05}, 0.2. 0.3, 1\} \\
                                    & Unlabeled batch size & \{128\} \\
        \midrule
        \multirow{2}{*}{CA-MAE (FM) + TVN} &  Learning rate & \{\textbf{0.001}, 0.002, 0.003, 0.004, 0.005, 0.006, 0.007, 0.009, 0.01\} \\
        & Batch size & \{128\}\\
        \midrule
        \multirow{2}{7em}{Style-ID} & Learning rate &                               \{0.001\} \\
                                    & Batch size  & \{128\} \\
        \midrule
        \multirow{2}{7em}{ARM-BN} & Learning rate & \{0.001\} \\
                                    & Batch size  & \{8, 16, 32, 64, \textbf{128}\} \\
        \midrule
        \multirow{2}{7em}{CS-ARM-BN} & Learning rate & \{0.001\} \\
                                    & Batch size  & \{8, 16, 32, \textbf{64}\} \\
                                    & Negative controls batch size & \{64, \textbf{128}\} \\
        \midrule
        \multirow{4}{7em}{ARM-CML} & Learning rate & \{0.001\} \\
                                    & Batch size  & \{128\} \\
                                    & Context channels  & \{12\} \\
                                    & Kernel size & \{5\} \\
        \bottomrule
    \end{tabular}
    }
\label{tab:hyperparams}
\end{table*}

\section{Dataset details}
\label{app:data}
As explained in Section \ref{sec:experiments}, the experiments were carried out on the JUMP-CP dataset, using source 3 and source 8. 
The classification task consists in classifying the MoA of the given image, similar to \citet{Haslum2023}. As we need images that are correctly annotated, we use images that were treated with compounds that have a clear and well-defined MoA. Specifically, we use the images treated with the compounds used as positive controls, which are all present in every plate. The name of these compounds, along with their mechanism of action can be found in Supplementary Table~\ref{tab:cpds}. Source 3 contains 9 experimental batches and consists of 49,491 microscopy images (with 5 channels each) and 172 different plates. Source 8 contains 4 experimental batches and 56,160 microscopy images (with 5 channels each) and 195 different plates. 
For experiments with the CellProfiler features, we download the provided images from the CellPainting Gallery \citep{CPgallery}
, and perform standardization by using the training features. We use these features as provided, and they were calculated on a well-level basis, so that this is the only baseline where accuracy is evaluated on a well level. 

\textit{Preprocessing.} As a preprocessing step, the images were downsized to a resolution of 512x512 and we performed illumination correction by clipping 0.01\% of the brightest pixels. During training, the images were further cropped to 256x256 and random vertical and horizontal flipping was performed. At inference time, the center crop of the images with a resolution of 256x256 was used. Moreover, in all methods, except for the baseline with instance norm and the MAE-based methods, the images were normalized using the mean and standard deviation calculated for the training split. In the remaining methods, the images are preprocessed by normalizing the channel of each image individually, in order to match the pretraining procedure in the case of the CA-MAE baselines \citep{Kraus2024}. 

For the TVN correction, additional negative controls, i.e. samples in which only DMSO was applied, were used. In total, the negative controls present in the 172 plates from the dataset described above, consist of 49,536 images. 

\begin{table*}[h!]
\centering
\caption{Compounds used as mechanism of action representatives for the classification task.}
\begin{tabular}{lll}
\toprule
\textbf{Compound name}                     & \textbf{Mechanism of action class}                              \\ \hline
Aloxistatin   &  Cysteine protease inhibitor                              \\
FK-866        & NAMPT inhibitor \\
AMG900        & Aurora kinase inhibitor                                  \\
Dexamethasone & Glucocorticoid agonist                                   \\
LY2109761     & TGF-$\beta$ inhibitor                                          \\
NVS-PAK1-1    & PAK1 inhibitor                                           \\
Quinidine     & Sodium channel inhibitor                                 \\
TC-S-7004    & DYRK1A/B inhibitor  \\                
\bottomrule
\end{tabular}
\label{tab:cpds}
\end{table*}

\section{Additional compared methods}
\label{appsec:methods}
\textbf{Domain-Adversarial Neural Networks (DANN)~\citep{ganin2014}.}
DANN uses a domain discriminator $\Bd_{\Bv}: \dR^D \to \mathcal S^{B-1}$ over 
features $\Ba=\Bf_{\Bw}(\Bx)$ and uses gradient reversal to promote domain invariance:
\begin{align}
\min_{\Bw} \max_{\Bv} \sum_{b=1}^B \sum_{(\Bx,\By)\in\mathcal D^{(b)}} 
\Big[ \mathcal L_{\mathrm{CE}}(\By, \Bg(\Bx,\Bw))
\;+\; \nonumber \\
\lambda \, \mathcal L_{\mathrm{CE}}\big(b, \Bd_{\Bv}(\Ba)  \big) \Big],
\end{align}
where the second loss term is cross-entropy on domain labels and $\lambda>0$ trades classification performance
against domain objectives.

\textbf{Correlation Alignment (CORAL)~\citep{sun2015}.}
Let $\BA_S=\{\Ba_i\}$ and $\BA_T=\{\Ba'_j\}$ be source and target features in a training batch; 
let $\BC_S$ and $\BC_T$ denote their covariance matrices. 
CORAL adds a moment-matching penalty to the supervised loss:
\begin{align}
\min_{\Bw}\; \sum_{b=1}^B\sum_{(\Bx,\By)\in \mathcal D^{(b)}} \mathcal \mathcal L_{\mathrm{CE}}(\By, \Bg(\Bx,\Bw))
\;+\; \nonumber \\
\gamma \,\| \BC_S - \BC_T\|_F^2,
\end{align}
aligning second moments of source and target features during training. We trained CORAL
on considering the samples in the training 
set to stem from the source distribution 
and the samples in the test set, from 
the target distribution.

\textbf{Channel-Agnostic MAE (CA-MAE) + Typical Variation Normalization (TVN)~\citep{Kraus2024,Ando2017}.}
As in the foundation model baseline, we use the embeddings 
from a frozen CA-MAE encoder, and perform a post-hoc correction with TVN. 
For TVN, given a set of batch-$\beta$ control embeddings 
from the target batch, $\{\Bz_j^{\beta}\}$, 
$\Bmu^{\mathrm{ctrl},\beta}$ and $\BSigma^{\mathrm{ctrl},\beta}$ 
are estimated
and used to apply a per-batch whitening:
\begin{align}
\tilde \Ba \;=\; \big(\BSigma^{\mathrm{ctrl},\beta}\big)^{-\frac{1}{2}}\;\big(\Ba - \Bmu^{\mathrm{ctrl},\beta}\big)
\end{align} 
is performed. Then, the covariance is aligned to 
the covariance of the training data $\BSigma^{\mathrm{train}}$.
The classifier $\Bc_{\theta}$ is then trained with the resulting re-aligned
representations $\tilde \Ba$. 
This matches first and second moments to the typical, 
i.e. control, distribution of each batch.

\textbf{Generative methods: StyleID \citep{Chung2024}.}
We used the diffusion-based style transfer method proposed by \citet{Chung2024} to convert all images to the style of one specific experimental batch
\citep{NIPS2016_502e4a16}. Then, as in the supervised baselines, a ResNet50 and a linear classifier are trained to minimize the cross-entropy loss $\mathcal L_{\mathrm{CE}}$ in the generated training samples. 

\textbf{DEYO \citep{Lee2024}.} DeYO is a test-time adaptation method that extends entropy-minimization approaches by selecting and weighting target samples according to whether their predictions appear to rely on semantically meaningful structure rather than spurious cues. In addition to entropy minimization, DeYO defines a pseudo-label probability difference (PLPD) score by comparing the confidence of the pseudo-label before and after applying a transformation $\mathcal T$ that destroys the semantic meaning:

\begin{align}
\mathrm{PLPD}(\Bx)
= ({\hat\By}_{\Bx}
-
 {\hat\By}_{\mathcal T(\Bx)})_{p},
\end{align}

where ${\hat\By}_{\Bx}$ and ${\hat\By}_{\mathcal T(\Bx)}$ are the model predictions of sample $\Bx$ and transformed sample $\mathcal T(\Bx)$, respectively, and $p = \argmax(\hat\By_{\Bx})$ is a pseudo-label estimated by the prediction of $\Bx$.

Samples with low entropy and high PLPD are selected and weighted more strongly during adaptation, so that the test-time objective can be written as:

\begin{align}
\min_{\kappa,\lambda} \phi(\Bx)\,\mathbb I[\Bx\in\mathcal S]\,
H({\hat\By}),
\end{align}

where $\mathcal S$ denotes the set of selected target samples and $\phi(\Bx)$ is a weighting function based on entropy and PLPD.
The BatchNorm  affine parameters are then updated by minimizing a weighted entropy over the selected target samples.

\textbf{ROID \citep{Marsden2023}.}
ROID is an online test-time adaptation method based on weighted self-training. For an unlabeled target image $\Bx$, ROID computes a certainty score from the prediction entropy and a diversity score relative to an exponential moving average of recent predictions, and combines them into a sample weight $w(\Bx)$. These weights are then used in a self-training objective so that adaptation focuses on target samples that are both confident and diverse. To improve stability, ROID additionally averages the adapted parameters with the source parameters and estimates the target class prior to correct for label shift.

\section{Additional experiments}
\label{appsec:experiments}

\subsection{Failure mode of BN adaptation --- dependence on adaptation batch size}
\label{appsec:exp3_batch_size}
BN-based test-time adaptation implicitly relies on accurate estimation of target statistics. When only a few samples are available from a new batch (or when adaptation must be performed in small micro-batches due to memory constraints), these estimates become noisy. To stress-test this failure mode, we vary the number of target samples used to compute adaptation statistics (\emph{adaptation batch size / context size} $M$) and evaluate methods that adapt BN parameters. Table~\ref{apptab:batch_size} reports the results.

\textbf{Previous BN-adaptation methods collapse for small $M$.}
Both AdaBN and TENT perform poorly when only very few target samples are available to estimate BN statistics (e.g., $M\in\{1,2,4\}$). Performance improves monotonically with $M$, approaching a plateau only once statistics can be reliably estimated (larger $M$). This explains why BN-based test-time adaptation can appear strong in benchmarks that provide a full target batch, yet fail in operational settings with limited or streaming access.

\textbf{Control-stabilized context mitigates the small-$M$ regime.}
Our control-stabilized variant (\textbf{CS-ARM-BN} in the table) remains strong even for extremely small numbers of perturbed samples. The key is that, while perturbed-class composition may be sparse, \emph{negative controls are plentiful and consistently available in every batch}. Using controls to estimate batch statistics yields a stable estimate of batch-specific nuisance variation, making adaptation substantially less sensitive to the number of available perturbed samples.

\begin{table*}[h]
\centering
\caption{Comparison of MoA classification accuracy of different methods in a new experimental batch with different batch sizes. The accuracy values denote mean and standard deviation over 5-fold cross-validation, thus five training re-runs of each method. }
\label{apptab:batch_size}

\resizebox{\linewidth}{!}{
\begin{tabular}{lccccccccc}
\toprule
\multirow{2}{*}{\textbf{Method}} & \multicolumn{8}{c}{\textbf{Labeled batch size}}     \\
                        & 1 & 2 & 4 & 8 & 16 & 32 & 64 & 128  & Full domain \\
\midrule
ERM (no adaptation)  &  0.862 ± 0.060 &  0.862 ± 0.060 & 0.862 ± 0.060  & 0.862 ± 0.060  &  0.862 ± 0.060  &  0.862 ± 0.060  &  0.862 ± 0.060  &  0.862 ± 0.060 & 0.862 ± 0.060 \\
TENT        &  0.126 ± 0.001 &  0.175 ± 0.013 & 0.230 ± 0.011  & 0.323 ± 0.012  &  0.550 ± 0.026  &  0.794 ± 0.021  &  0.881 ± 0.025  &  0.889 ± 0.031 & 0.908 ± 0.032 \\
AdaBN & 0.130 ± 0.005 & 0.447 ± 0.095 & 0.709 ± 0.052 & 0.839 ± 0.024 & 0.898 ± 0.014 & 0.919 ± 0.014 & 0.924 ± 0.014 & 0.927 ± 0.013 & 0.928 ± 0.013 \\
ARM-BN & 0.126 ± 0.007 & 0.514 ± 0.092 & 0.716 ± 0.052 & 0.832 ± 0.028 & 0.896 ± 0.019 & 0.923 ± 0.018 & \textbf{0.930 ± 0.018} & \textbf{0.933 ± 0.017} & \textbf{0.935 ± 0.018} \\
CS-ARM-BN & \textbf{0.923 ± 0.018} & \textbf{0.923 ± 0.018} & \textbf{0.924 ± 0.019} & \textbf{0.925 ± 0.018} & \textbf{0.927 ± 0.019} & \textbf{0.929 ± 0.019} & \textbf{0.930 ± 0.019} & 0.929 ± 0.020 & 0.923 ± 0.023 \\
\bottomrule
\end{tabular}
}
\end{table*}

\subsubsection{Ablation study. Effect of the negative controls set sample size}
In the previous experiment, we assume that other methods do not consider the presence of negative controls in biological data, so that the available samples are only perturbed ones, while in CS-ARM-BN we make use of negative controls, which results in a larger effective batch size. However, in this experiment we also study the effectiveness of standard ARM-BN in the presence of negative controls, so that we compare both methods using the same total batch size. In Table~\ref{apptab:batch_size_ablation}, we show that \textbf{CS-ARM-BN presents a higher accuracy} even when the \textbf{same amount of unlabeled data} is available, probably due to the fact that during the meta-learning phase, the model has already learned to normalize the intermediate features of the model using both negative controls and perturbed samples. 

\begin{table}[h]
\centering
\caption{Comparison between ARM-BN and CS-ARM-BN classification accuracy using the same number of samples in every batch, with the same proportion of negative controls per perturbed sample. The accuracy values denote mean and standard deviation over 5-fold cross-validation, thus five training re-runs of each method.}
\label{apptab:batch_size_ablation}
\begin{tabular}{lccccc}
\toprule
\multirow{2}{*}{\textbf{Method}} & \multicolumn{4}{c}{\textbf{Total batch size [perturbed : controls]}} \\
& 5 [1 : 4] & 10 [2 : 8] & 20 [4 : 16] & 40 [8 : 32] & 80 [16 : 64]\\
\midrule
ARM-BN            & 0.799 ± 0.020 & 0.827 ± 0.024 & 0.846 ± 0.025 & 0.860 ± 0.026 & 0.870 ± 0.028 \\
CS-ARM-BN (ours)  & \textbf{0.848 ± 0.035} & \textbf{0.875 ± 0.029} & \textbf{0.891 ± 0.028} & \textbf{0.899 ± 0.027} & \textbf{0.905 ± 0.026}\\
\bottomrule
\end{tabular}

\end{table}

\newpage
\subsection{Label shift experiment. Supplementary information.}
\label{appsec:label_shift}
\subsubsection{Experiment details}
In experiment \ref{sec:exp3_label_shift} we show a comparison between methods under label shifts. To achieve batches that have a certain level of label shift, we sample a class probability vector from a Dirichlet distribution with a certain $\alpha$ parameter. Then we use this class probability vector to define a multinomial distribution and draw N samples from it, that we will use as input to our model. In total we use a labeled batch size of 36 in every method, and we include 288 negative samples in CS-ARM-BN. 

One could argue that performance differences might be only due to a larger sample size, therefore, we show in Table \ref{apptab:label_shift_ablation} other baselines including the same amount of negative controls in each batch, as explained in the following section.

\subsubsection{Ablation study. Effect of the negative controls set sample size}
 In our method, CS-ARM-BN, negative controls are used both at training and at inference time, so the total batch size used to compute the BN parameters is larger than for the other methods. To demonstrate that the gains in performance in our method are not only due to a larger batch size but also due to the fact that the models were trained in a meta-learning fashion, we run the \textbf{AdaBN and ARM-BN} with the same amount of negative controls and \textbf{same total batch size as for CS-ARM-BN}. We don't perform this experiment for TENT because we argue that there is no reason why minimizing the entropy of the model outputs for control samples should be beneficial. In Table~\ref{apptab:label_shift_ablation} we show that, even if including controls increases the performance of both methods, \textbf{CS-ARM-BN still shows} \textbf{the best performance}.

\begin{table}[h]
\centering
\caption{Comparison of MoA classification accuracy of different methods in a new experimental batch under different degrees of label shift. A smaller $\alpha$ value corresponds to a larger degree of label shift.}
\label{apptab:label_shift_ablation}
\begin{tabular}{lcccc}
\toprule
\multirow{2}{*}{\textbf{Method}} & \multicolumn{4}{c}{\textbf{Label shift degree}}     \\
                          & $\alpha=4$  & $\alpha=1$ & $\alpha=0.1$ & $\alpha=0.01$ \\
\midrule
ERM (no adaptation)     & 0.862 ± 0.060 & 0.862 ± 0.060 & 0.862 ± 0.061  &  0.862 ± 0.061 \\
\midrule
TENT                    & 0.887 ± 0.014 & 0.832 ± 0.014 & 0.778 ± 0.016 & 0.156 ± 0.010 \\
DEYO                    & 0.887 ± 0.013 & 0.830 ± 0.014 & 0.773 ± 0.015 & 0.156 ± 0.010 \\
ROID                    & 0.894 ± 0.016 & 0.837 ± 0.017 & 0.792 ± 0.020 & 0.157 ± 0.011 \\
Ada-BN                  & 0.899 ± 0.012 & 0.851 ± 0.016 & 0.452 ± 0.010 & 0.229 ± 0.009\\
ARM-BN                  & 0.894 ± 0.018 & 0.831 ± 0.017 & 0.416 ± 0.020 & 0.228 ± 0.009 \\
Ada-BN (perturbed + controls)       &\textit{ 0.879 ± 0.018} & \textit{0.871 ± 0.019} & \textit{0.815 ± 0.022} & \textit{0.785 ± 0.033}\\
ARM-BN  (perturbed + controls)     & 0.861 ± 0.032 & 0.850 ± 0.033 & 0.785 ± 0.058 & 0.752 ± 0.068 \\
CS-ARM-BN (ours)        & \textbf{0.927 ± 0.017} &\textbf{0.924 ± 0.018} & \textbf{0.906 ± 0.021} & \textbf{0.894 ± 0.022} \\
\bottomrule
\end{tabular}

\end{table}

\newpage

\subsection{Evaluation of CS-ARM-BN with a Vision Transformer-based architecture}
Although, throughout the main manuscript, we have evaluated our method and the baselines using a ResNet50, in this section we also benchmark the best performing methods using LeViT \citep{Graham2021}, an architecture based on the Vision Transformer that employs Batch Normalization layers instead of the usual Layer Normalization. Particularly, we evaluate the methods on the most challenging task, under both label shift and strong domain shift.  In Table~\ref{tab:label_shift_levit}, it can be seen that adapting the Batch Normalization layers and including negative controls in a meta-learning fashion is an effective strategy to mitigate domain and label shifts also when using a Vision Transformer architecture. We evaluated ERM, TENT, AdaBN and ARM-BN as in previous experiments, and additionally included two more recent TTA baselines, DEYO \citep{Lee2024} and ROID \citep{Marsden2023}.

\begin{table*}[h]
\centering
\caption{Comparison of MoA classification accuracy of different methods across target sources under different degrees of label shift, using a \textbf{LeViT, a vision transformer-based architecture}. A lower $\alpha$ denotes a higher degree of label shift. \textbf{The accuracy values denote mean and standard deviation over re-runs with three different seeds.}}
\vspace{0.2em}
\label{tab:label_shift_levit}
\begin{threeparttable}
\begin{tabular}{l cc cc}
\toprule
\multirow{2}{*}{\textbf{Method}} &
\multicolumn{2}{c}{\textbf{S3 $\rightarrow$ S8}} &
\multicolumn{2}{c}{\textbf{S8 $\rightarrow$ S3}} \\
 \cmidrule(lr){2-3} \cmidrule(lr){4-5}
 & $\alpha$ = 1 & $\alpha$ = 0.01  
 & $\alpha$ = 1 & $\alpha$ = 0.01   \\
\midrule

(LeViT) no adaptation (BatchNorm) 
& 0.607±0.023 & 0.608±0.019
& 0.536±0.062 & 0.536±0.051\\

\midrule
(LeViT) TENT 
& 0.675±0.017 & 0.137±0.002
& 0.640±0.013 & 0.145±0.005 \\

(LeViT) DEYO 
& 0.674±0.014 & 0.137±0.002 
& 0.639±0.015 & 0.145±0.005 \\

(LeViT) ROID 
& 0.694±0.018 &  0.138±0.003
& 0.657±0.015 &  0.147±0.005 \\

(LeViT) AdaBN 
& 0.699±0.015 & 0.165±0.002 
& 0.665±0.014 & \textit{0.168±0.004}\\

(LeViT) ARM-BN  
& \textit{0.735±0.024} & \textit{0.166±0.003}
& \textit{0.702±0.119} & 0.165±0.001\\

(LeViT) CS-ARM-BN (ours)
& \textbf{0.791±0.025} & \textbf{0.707±0.021}
& \textbf{0.722±0.015} & \textbf{0.610±0.017} \\

\bottomrule
\end{tabular}

\end{threeparttable}
\end{table*}

\subsection{Microscopy imaging data structure and domain alignment}
\label{appsec:micro_structure}

Microscopy imaging data presents a hierarchical structure. In each imaging platform, also called \textit{source} throughout this manuscript, reagents and cell cultures change over time and have to be replaced, which leads to different \textit{experimental batches}. Moreover, within each one of these experimental batches, cells have to be laid out in what are known as \textit{plates}, which are then placed under the microscope and images are obtained. As we have explained in Section \ref{sec:intro}, there are visible differences between plates, batches and sources. Moreover, each one of these plates has different positions, known as \textit{wells}. 

Given this hierarchy, a natural question is at which level to perform the domain alignment. While each source or batch has a larger number of samples, they are a set of other domains which might lead to a non-optimal alignment. In the following experiment, we show the adaptation results for AdaBN at the possible different levels: source, plate and batch, and we show that aligning at the plate level yields better results, even if fewer samples are available per domain. 
\begin{table}[h]
\centering
\caption{MoA classification accuracy of the AdaBN method adapting the BatchNorm statistics per source, per batch and per plate. Even if each plate consists of, naturally, fewer samples than each batch or source, the resulting accuracy is higher when the domain adaptation is performed per plate. Accuracy is an average over the S3$\rightarrow$S8 and S8$\rightarrow$S3 transfer experiments, shown in Table \ref{tab:generalization_full}.}
\label{apptab:adabn_alignment}
\begin{tabular}{lcccc}
\toprule
\textbf{Alignment level} &  \textbf{Accuracy}  & \textbf{Number of domains} &  \textbf{Samples per domain (avg.)} \\
\midrule
Source                    & 0.773 ± 0.033 & 1   &   56,160  \\
Batch                     & 0.790 ± 0.014 & 4   &  14,040     \\
Plate                     & 0.798 ± 0.014 & 195 &  288\\
\bottomrule
\end{tabular}

\end{table}
\newpage
\section{The in-context learning view on CS-ARM-BN}

Our method 
CS-ARM-BN can be understood as a form of \emph{in-context 
learning}: rather than retraining the model, the classifier 
conditions its predictions on a small \emph{context} of 
unlabeled samples from the new experimental batch, without 
any gradient update. The negative controls play the role of 
a stable, class-agnostic reference that directly reflects 
the batch-specific technical variation.

As shown in Algorithm~\ref{alg:cs-arm-bn-detail}, this 
can be written in two equivalent ways: either as an explicit 
two-step procedure of calibration followed by inference 
(\texttt{model.adapt}), or as a single context-conditioned 
forward pass (\texttt{model.cpredict}), making the analogy 
to in-context learning explicit.
\begin{center}
\begin{minipage}{0.51\linewidth}
\begin{algorithm}[H]
\caption{Pseudo code for 
CS-ARM-BN vs traditional image classifier
pipelines | In-context view on CS-ARM-BN}
\label{alg:cs-arm-bn-detail}
\begin{lstlisting}[language=Python, 
                   basicstyle=\footnotesize\ttfamily,
                   commentstyle=\color{jku_blue},
                   keywordstyle=\color{black},
                   showstringspaces=false,
                   breaklines=true,
                   breakatwhitespace=true,
                   frame=none,
                   aboveskip=0pt,
                   belowskip=0pt,
                   escapechar=~]
### Standard Image Classifier Pipeline ###
# no calibration of classifier to data
x = load_data() # your biomedical images
model = load_model() # public classifier
pred = model.predict(x)  # domain gap!

### CS-ARM-BN Classifier Pipeline ###
# adapt/calibrate classifier to your data
x = load_data() # your biomedical images
z = load_controls() # available controls
model = load_model() # public classifier
~{\color{jku_red}\bfseries model.adapt(concat([x, z]))}~  # calibrate
pred = model.predict(x)  # neutralized gap

### In-context learning view on CS-ARM-BN
# adapt/calibrate classifier to your data
x = load_data() # your biomedical images
z = load_controls() # available controls
model = load_model() # public classifier
context = concat([x, z]) # assemble context
~{\color{jku_red}\bfseries pred = model.cpredict(x, context=context) }~  
\end{lstlisting}
\end{algorithm}
\end{minipage}
\end{center}

\section{Details on estimators}
\label{sec:details-estimators}
\subsection{Setup}

We model the BN activation mean for a sample $x$ as:
\begin{align}
    \mu_{\mathrm{obs},x} = \mu_{\mathrm{domain}} + \mu_{\mathrm{class}(x)} + \varepsilon_x,
\end{align}
where $\mu_{\mathrm{domain}}$ is the batch-specific technical offset we wish to estimate,
$\mu_{\mathrm{class}(x)} = 0$ for control samples,
$\mu_{\mathrm{class}(x)} \neq 0$ for perturbed samples,
and $\varepsilon_x \sim (0, \sigma^2)$ is additive noise.

We have $C$ control samples and $L$ perturbed samples, with $M = C + L$ total.
The mean class effect across all perturbed samples is
\begin{align}
    \bar{\mu}_{\mathrm{class}} = \frac{1}{L}\sum_{l=1}^{L} \mu_{\mathrm{class}(l)}.
\end{align}
For any estimator $\hat{\mu}$, the MSE decomposes as:
\begin{align}
    \mathrm{MSE}(\hat{\mu}) = \mathrm{Bias}^2(\hat{\mu}) + \mathrm{Var}(\hat{\mu}).
\end{align}

\subsection{Estimator 1: ARM-BN and AdaBN (Perturbed Samples Only)}

The estimator is the sample mean over the $L$ perturbed samples:
\begin{align}
    \hat{\mu}^{\mathrm{ARM-BN}}
    = \frac{1}{L}\sum_{l=1}^{L} \mu_{\mathrm{obs},l}
    = \frac{1}{L}\sum_{l=1}^{L}
      (\mu_{\mathrm{domain}} + \mu_{\mathrm{class}(l)} + \varepsilon_l).
\end{align}

\paragraph{Bias.}
\begin{align}
    \mathbb{E}[\hat{\mu}^{\mathrm{ARM-BN}}]
    &= \mu_{\mathrm{domain}} + \bar{\mu}_{\mathrm{class}}, \\
    \mathrm{Bias}
    &= \mathbb{E}[\hat{\mu}^{\mathrm{ARM-BN}}] - \mu_{\mathrm{domain}}
     = \bar{\mu}_{\mathrm{class}}.
\end{align}
The bias equals the mean class effect, which grows large under label shift.

\paragraph{Variance.}
\begin{align}
    \mathrm{Var}(\hat{\mu}^{\mathrm{ARM-BN}})
    = \mathrm{Var}\left(\frac{1}{L}\sum_{l=1}^{L}\varepsilon_l\right)
    = \frac{\sigma^2}{L}.
\end{align}

\paragraph{MSE.}
\begin{align}
    \mathrm{MSE}(\hat{\mu}^{\mathrm{ARM-BN}})
    = \bar{\mu}_{\mathrm{class}}^{2} + \frac{\sigma^2}{L}.
\end{align}

\subsection{Estimator 2: Controls Only}

The estimator is the sample mean over the $C$ control samples.
Since controls satisfy $\mu_{\mathrm{class}} = 0$:
\begin{align}
    \hat{\mu}^{\mathrm{ctrl}}
    = \frac{1}{C}\sum_{c=1}^{C} \mu_{\mathrm{obs},c}
    = \frac{1}{C}\sum_{c=1}^{C}
      (\mu_{\mathrm{domain}} + \varepsilon_c).
\end{align}

\paragraph{Bias.}
\begin{align}
    \mathbb{E}[\hat{\mu}^{\mathrm{ctrl}}]
    &= \mu_{\mathrm{domain}}, \\
    \mathrm{Bias} &= 0.
\end{align}
The estimator is unbiased because controls carry no class-specific signal by definition.

\paragraph{Variance.}
\begin{align}
    \mathrm{Var}(\hat{\mu}^{\mathrm{ctrl}}) = \frac{\sigma^2}{C}.
\end{align}
This is large when $C$ is small, which is the typical regime in drug-screening experiments.

\paragraph{MSE.}
\begin{align}
    \mathrm{MSE}(\hat{\mu}^{\mathrm{ctrl}})
    = 0 + \frac{\sigma^2}{C}.
\end{align}

\subsection{Estimator 3: CS-ARM-BN (Controls + Perturbed Samples)}

The estimator is the sample mean over all $M = C + L$ samples:
\begin{align}
    \hat{\mu}^{\mathrm{CS-ARM-BN}}
    &= \frac{1}{M}\sum_{u \in \mathcal{C}^{\beta}} \mu_{\mathrm{obs},u}
     = \frac{1}{M}
       \left(
         \sum_{c=1}^{C}\mu_{\mathrm{obs},c}
         + \sum_{l=1}^{L}\mu_{\mathrm{obs},l}
       \right).
\end{align}
Expanding each term:
\begin{align}
    \hat{\mu}^{\mathrm{CS-ARM-BN}}
    &= \frac{1}{M}
       \left(
         \sum_{c=1}^{C} (\mu_{\mathrm{domain}} + \varepsilon_c)
         + \sum_{l=1}^{L} (\mu_{\mathrm{domain}} + \mu_{\mathrm{class}(l)} + \varepsilon_l)
       \right) \nonumber \\
    &= \mu_{\mathrm{domain}}
       + \frac{1}{M}\sum_{l=1}^{L}\mu_{\mathrm{class}(l)}
       + \frac{1}{M}\sum_{u}\varepsilon_u \nonumber \\
    &= \mu_{\mathrm{domain}}
       + \frac{L}{M} \bar{\mu}_{\mathrm{class}}
       + \frac{1}{M}\sum_{u}\varepsilon_u.
\end{align}

\paragraph{Bias.}
\begin{align}
    \mathbb{E}[\hat{\mu}^{\mathrm{CS-ARM-BN}}]
    &= \mu_{\mathrm{domain}} + \frac{L}{M} \bar{\mu}_{\mathrm{class}}, \\
    \mathrm{Bias}
    &= \frac{L}{M} \bar{\mu}_{\mathrm{class}}.
\end{align}
The bias is attenuated by the factor $L/M < 1$ relative to Estimator~1.
The more controls are included (larger $C$, smaller $L/M$), the smaller the bias.

\paragraph{Variance.}
\begin{align}
    \mathrm{Var}(\hat{\mu}^{\mathrm{CS-ARM-BN}})
    = \mathrm{Var}\left(\frac{1}{M}\sum_{u}\varepsilon_u\right)
    = \frac{\sigma^2}{M}.
\end{align}
Since $M > L$ and $M > C$, this variance is strictly smaller than
those of Estimators~1 and~2.

\paragraph{MSE.}
\begin{align}
    \mathrm{MSE}(\hat{\mu}^{\mathrm{CS-ARM-BN}})
    = \left(\frac{L}{M}\right)^{2} \bar{\mu}_{\mathrm{class}}^{2}
    + \frac{\sigma^2}{M}.
\end{align}

\subsection{Comparison}

\begin{table}[h]
\centering
\caption{MSE decomposition for the three estimators of $\mu_{\mathrm{domain}}$.}
\label{tab:mse_comparison}
\begin{tabular}{lccc}
\hline
\textbf{Estimator} & \textbf{Bias\textsuperscript{2}} & \textbf{Variance} & \textbf{Notes} \\
\hline
ARM-BN and AdaBN & $\bar{\mu}_{\mathrm{class}}^{2}$             & $\sigma^2/L$ & Large bias under label shift \\
Controls-only & $0$                                         & $\sigma^2/C$ & Large variance when $C$ is small \\
CS-ARM-BN & $(L/M)^{2} \bar{\mu}_{\mathrm{class}}^{2}$  & $\sigma^2/M$ & Trade-off between the two above \\
\hline
\end{tabular}
\end{table}

CS-ARM-BN achieves the lowest MSE by trading a residual bias
shrunk by $(L/M)^2$ relative to ARM-BN against the lowest variance
of all three estimators (since $M \geq \max(L,C)$).
This bias-variance trade-off is most favourable in the typical
drug-screening regime: few controls ($C$ small) and
moderate-to-severe label shift.

\section{Detecting edge cases with uncertainty}

\begin{figure}
    \centering
    \includegraphics[width=0.8\textwidth]{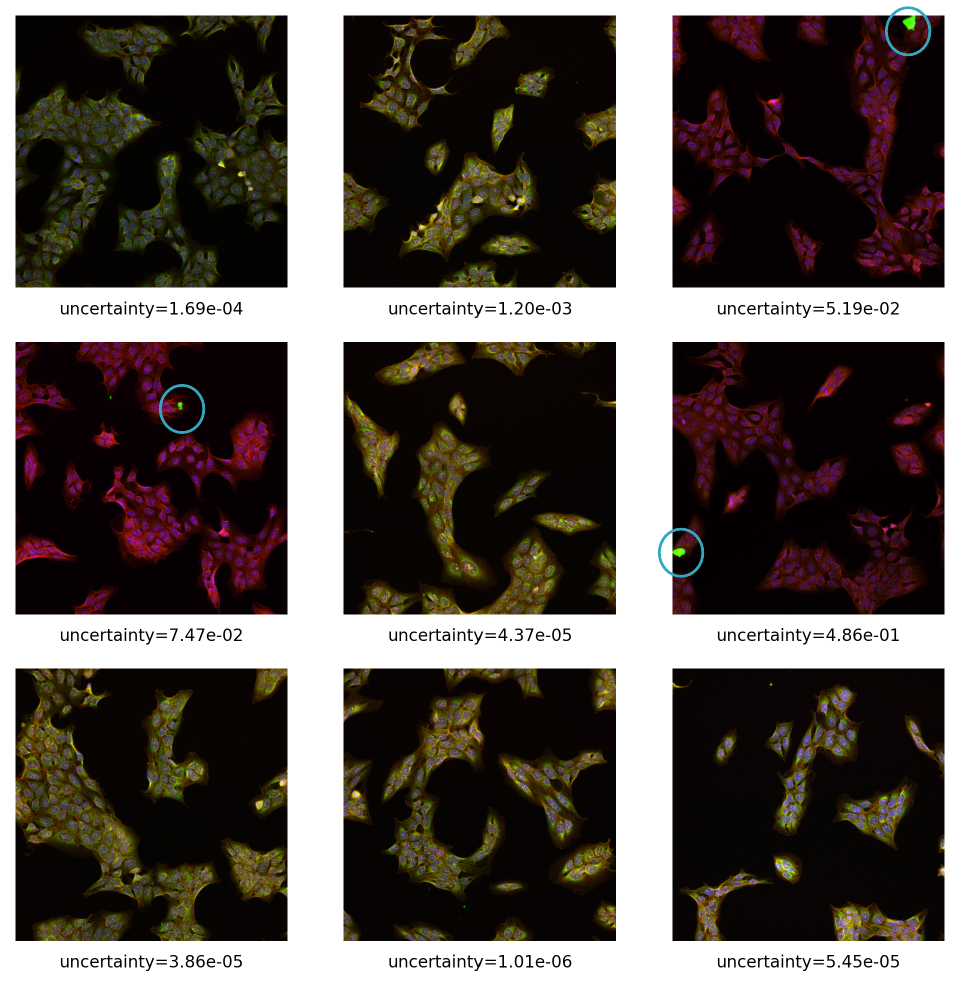}
     \caption{Representative examples of model predictions with corresponding deep ensemble uncertainty scores. Most samples exhibit low uncertainty (from approximately $1 \times 10^{-6}$ to $1 \times 10^{-3}$), reflecting stable predictions. In contrast, a subset of images shows markedly higher uncertainty (up to $4.86 \times 10^{-1}$), often associated with atypical visual artifacts such as localized high-intensity pixels in the green channel (highlighted with blue circles). These results suggest that such artifacts correspond to out-of-distribution, sample-specific effects that increase uncertainty.}
     \label{appfig:uncertainty}
\end{figure}

To further analyze whether the model can identify atypical samples at inference time, we performed an additional experiment based on predictive uncertainty. In our setup, the cross-validation folds and repeated training runs naturally provide a deep ensemble, which we use to estimate uncertainty for each test image. Specifically, we compute the entropy of the mean predicted class probabilities across all ensemble members. We then inspect samples with different uncertainty levels to determine whether highly uncertain predictions correspond to potential edge cases. As shown in Figure~\ref{appfig:uncertainty}, most images exhibit very low uncertainty values (approximately $10^{-6}$ to $10^{-3}$), indicating stable predictions. In contrast, a subset of samples shows substantially higher uncertainty, reaching up to approximately $10^{-1}$. Visual inspection reveals that these samples often contain unusual technical artifacts, such as extreme high-intensity pixels in the green channel, suggesting that uncertainty can serve as a useful signal for detecting sample-specific, non-biological artifacts that are not well represented in the training data.

\section{Test-Time Scaling}
\label{appsec:test_time_scaling}
\begin{figure}[t]
\centering

\begin{subfigure}[t]{0.49\textwidth}
    \centering
    \vspace{0pt}
    \includegraphics[height=5.5cm]{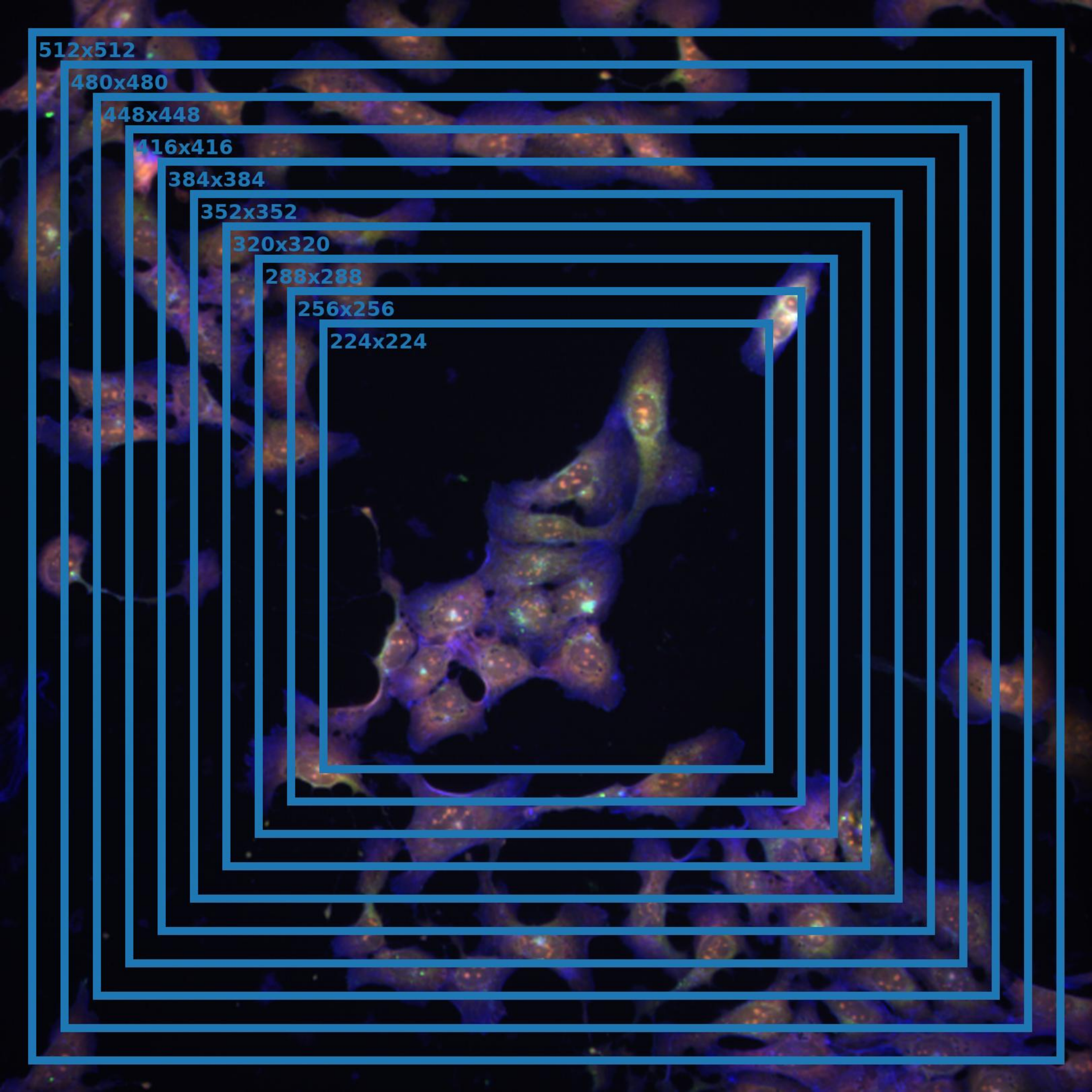}
    \caption{}
    \label{fig:first}
\end{subfigure}%
\hspace{-0.05\textwidth}
\begin{subfigure}[t]{0.49\textwidth}
    \centering
    \vspace{6pt}
    \includegraphics[height=5.5cm]{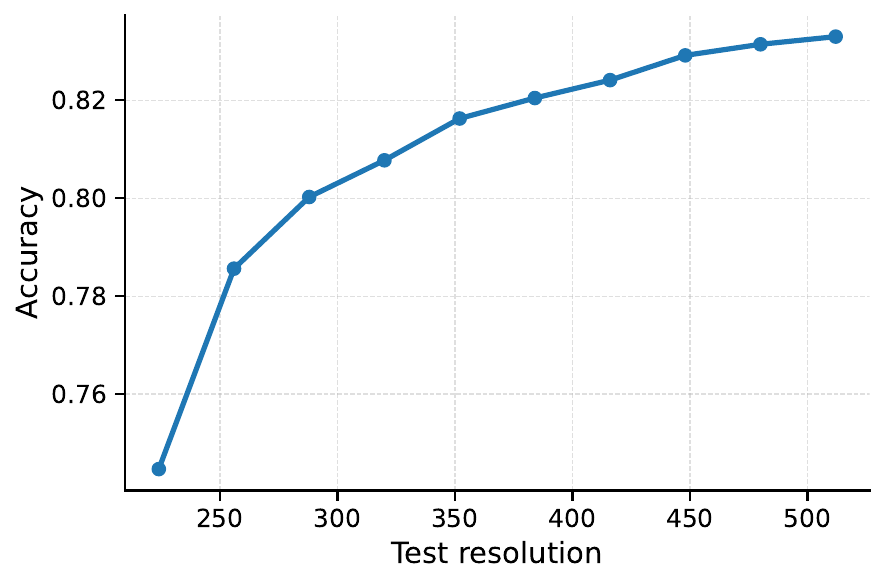}
    \caption{}
    \label{fig:second}
\end{subfigure}

\caption{
Test-time scaling for MoA prediction.
\textit{(a)} Example microscopy image center-cropped at increasing input resolutions, illustrating how higher-resolution crops retain additional spatial context and visual detail.
\textit{(b)} MoA classification accuracy as a function of test-time input resolution, with the trained model kept fixed. Accuracy improves as the inference resolution increases, suggesting that the learned representations can benefit from higher-resolution inputs without retraining.
}
\label{appfig:tts}
\end{figure}

In this section, we evaluate the effect of varying image resolution at inference time while keeping the trained model fixed. For computational efficiency, in the experiments reported in the main paper, we use a center crop of each image at inference time, as described in Section~\ref{app:data}. However, this protocol may discard useful visual information that is present outside the cropped region or only becomes available when the image is processed at a higher spatial resolution. As shown in Figure~\ref{appfig:tts}, increasing the input resolution at test time leads to improved performance, even though the model itself is not retrained under this modified setting. This suggests that the learned representations can benefit from higher resolution at inference time. These results highlight a trade-off between computational cost and predictive performance, and indicate that further gains may be obtained by using higher-resolution inputs when inference-time efficiency is less constrained.

\end{document}